\documentclass[runningheads]{llncs}
\usepackage{graphicx}
\usepackage{amsmath,amssymb} % define this before the line numbering.
\usepackage{color}

\usepackage{graphicx}
\usepackage{subcaption}
\graphicspath{{./figures/}}
\usepackage{amsmath}
\usepackage{amssymb}
\usepackage{booktabs}

\usepackage{algorithm, setspace}
\usepackage{algpseudocode, array, multirow}
\usepackage{caption}
\usepackage{xcolor}
\usepackage{extarrows}
\usepackage{enumitem}
\usepackage{booktabs}
\usepackage{rotating}
\usepackage{fixltx2e}
\usepackage{textcomp}

\definecolor{darkgreen}{rgb}{0.035, 0.412, 0.098}
\definecolor{edits}{rgb}{0.0, 0.0, 0.0}
\usepackage[breaklinks,colorlinks]{hyperref}
\hypersetup{colorlinks,breaklinks,urlcolor=[rgb]{0,0,0},linkcolor=[rgb]{0,0,0},citecolor=[rgb]{0,0,0}}
\usepackage{orcidlink}

\usepackage[capitalize]{cleveref}
\crefname{section}{Sec.}{Secs.}
\crefname{table}{Tab.}{Tabs.}
\Crefname{equation}{Eq.}{Eqs.}
\Crefname{figure}{Fig.}{Figs.}

\newcommand{\etal}{\textit{et al.}}

\DeclareMathOperator*{\argmax}{arg\,max}
\DeclareMathOperator*{\argmin}{arg\,min}

\begin{document}
\pagestyle{headings}
\mainmatter

\def\ACCV22SubNumber{768}  % Insert your submission number here

%===========================================================
\title{\textsc{COLLIDER}: A Robust Training Framework for Backdoor Data} % Replace with your title

\titlerunning{\textsc{COLLIDER}: A Robust Training Framework for Backdoor Data}
\author{Hadi M.~Dolatabadi\textsuperscript{\orcidlink{0000-0001-9418-1487}} \and
Sarah Erfani\textsuperscript{\orcidlink{0000-0003-0885-0643}} \and
Christopher Leckie\textsuperscript{\orcidlink{0000-0002-4388-0517}}}
\authorrunning{H.~M.~Dolatabadi~\etal}
\institute{School of Computing and Information Systems\\The University of Melbourne\\Parkville, Victoria, Australia\\
\email{hadi.mohagheghdolatabadi@student.unimelb.edu.au}}
\maketitle

%===========================================================
\begin{abstract}
    Deep neural network~(DNN) classifiers are vulnerable to backdoor attacks.
	An adversary poisons some of the training data in such attacks by installing a trigger.
	The goal is to make the trained DNN output the attacker's desired class whenever the trigger is activated while performing as usual for clean data.
	Various approaches have recently been proposed to detect malicious backdoored DNNs.
	However, a robust, end-to-end training approach, like adversarial training, is yet to be discovered for backdoor poisoned data.
	In this paper, we take the first step toward such methods by developing a robust training framework, \textsc{Collider}, that selects the most prominent samples by exploiting the underlying geometric structures of the data. Specifically, we effectively filter out candidate poisoned data at each training epoch by solving a geometrical coreset selection objective.
	We first argue how clean data samples exhibit (1) gradients similar to the clean majority of data and (2) low local intrinsic dimensionality~(LID).
	Based on these criteria, we define a novel coreset selection objective to find such samples, which are used for training a DNN.
	We show the effectiveness of the proposed method for robust training of DNNs on various poisoned datasets, reducing the backdoor success rate significantly.
	
\end{abstract}

%===========================================================
\section{Introduction}\label{sec:introduction}
Deep neural networks~(DNN) have gained unprecedented attention recently and achieved human-level performance in various tasks such as object detection~\cite{russakovsky2015imagenet}.
Due to their widespread success, neural networks have become a promising candidate for use in safety-critical applications, including autonomous driving~\cite{geiger2012areweready,lillicrap2016continuous} and face recognition~\cite{deng2019arcface,schroff2015facenet}.
Unfortunately, it has been shown that neural networks may exhibit unexpected behavior when facing an adversary.

It has been shown that neural networks suffer from \textit{backdoor attacks}~\cite{gu2017badnets}. 
In such attacks, the attacker has control over the training process.
Usually, the adversary poisons a portion of the training data by installing a trigger on natural inputs.
Then, the neural network is trained either by the adversary or the user on this poisoned training data.
The attacker may add its trigger to the inputs at inference time to achieve its desired output.
However, such poisoned neural networks behave ordinarily on clean data. 
As such, defending neural networks against backdoor attacks can empirically be an arduous task.

In the most common setting for backdoor defense, motivated by the rise of Machine Learning as a Service (MLaaS)~\cite{shokri2017membership}, it is assumed that the user outsources training of its desired model to a third party.
The adversary then can exploit this freedom and provide a malicious, backdoored neural network to the user~\cite{liu2018finepruning,wang2019neuralcleanse,chen2019deepinspect,kolouri2020ulp,sikka2020detecting}.
From this perspective, the status-quo defense strategies against backdoor attacks can be divided into two categories~\cite{goldlum2020dataset}.
In \textit{detection-based} methods the goal is to identify maliciously trained neural networks~\cite{wang2020practical,kolouri2020ulp}.
\textit{Erasing-based} approaches, in contrast, try to effectively eliminate the backdoor data ramifications in the trained model, and hence, give a pure, backdoor-free network~\cite{liu2018finepruning,wang2019neuralcleanse,liu2020removing,zhao2020bridging,yige2021nad}.

A less-explored yet realistic scenario in defending neural networks against backdoor poisonings is when the user obtains its training data from untrustworthy sources.
Here, the attacker can introduce backdoors into the user's model by solely poisoning the training data~\cite{gu2017badnets,tran2018spectral,goldlum2020dataset,hayase2021defense}.
In this setting, existing approaches have several disadvantages.
First, these methods may require having access to a \textit{clean held-out validation dataset}~\cite{tran2018spectral}.
This assumption may not be valid in real-world applications where collecting new, reliable data is costly.
Moreover, such approaches may need a \textit{two-step training procedure}: a neural network is first trained on the poisoned data.
Then, the backdoor data is removed from the training set using the previously trained network.
After purification of the training set, the neural network needs to be re-trained~\cite{tran2018spectral,hayase2021defense}.
Finally, some methods achieve robustness by \textit{training multiple neural networks} on subsets of training data to enable a ``majority-vote mechanism''~\cite{goldlum2020dataset,levine2021deep,jia2020intrinsic}.
These last two assumptions may also prove expensive in real-world applications where it is more efficient to train a \textit{single} neural network only \textit{once}.
As a result, one can see that a standard, robust, and end-to-end training approach, like adversarial training, is still lacking for training on backdoor poisoned data.

To address these pitfalls, in this paper we leverage the theory of coreset selection~\cite{harpeled2004oncoresets,agarwal2004approx,campbell2018coreset,mirzasoleiman2020craig,mirzasoleiman2020crust} for end-to-end training of neural networks.
In particular, we aim to sanitize the possibly malicious training data by training the neural network on a subset of the training data.
To find this subset in an online fashion, we exploit coreset selection by identifying the properties of the poisoned data.
To formulate our coreset selection objective, we argue that the \textit{gradient space characteristics} and \textit{local intrinsic dimensionality}~(LID) of poisoned and clean data samples are different from one another.
We empirically validate these properties using various case studies.
Then, based on these two properties, we define an appropriate coreset selection objective and effectively filter out poisoned data samples from the training set.
As we shall see, this process is done online as the neural network is being trained.
As such, we can effectively eliminate the previous methods' re-training requirement.
We empirically show the successful performance of our method, named \textsc{Collider}, in training robust neural networks under various backdoor data poisonings resulting in about $25$\% faster training.

Our contributions can be summarized as follows:
\begin{itemize}
	\item To the best of our knowledge, we are the first to introduce a practical algorithm for single-run training of neural networks on backdoor data by using the idea of coreset selection. 
	\item We characterize clean data samples based on their gradient space and local intrinsic dimensionality and define a novel coreset selection objective that effectively selects them.  
	\item We perform extensive experiments under different settings to show the excellent performance of the proposed approach in reducing the effect of backdoor data poisonings on neural networks in an online fashion. 
\end{itemize}

%===========================================================
\section{Related Work}\label{sec:related_work}

This section reviews some of the most related work to our proposed approach.
For a more thorough overview of backdoor attacks and defense, please see~\cite{li2020survey,goldlum2020dataset}

%------------------------------------------------------------------------- 
\subsection{Backdoor Attacks}
In BadNets, Gu~\etal~\cite{gu2017badnets} showed that neural network image classifiers suffer from backdoor attacks for the first time. 
Specifically, the training data is poisoned by installing small triggers in the shape of single pixels or checkerboard patterns on a few images.
This poisoned data may come from any class.
Therefore, their label needs to be modified to the target class by the adversary.
As the labels are manipulated in addition to the training data, this type of backdoor data poisoning is known as \textit{dirty-label} attacks.
Similar findings have also been demonstrated on face-recognition networks using dirty-label data poisoning~\cite{chen2017targeted,liu2018trojaning}.

Different from dirty-label attacks, one can effectively poison the training data without changing the labels.
As the adversary does not alter the labels even for the poisoned data, such attacks are called \textit{clean-label}~\cite{shafahi2018frogs,turner2019label-consistent}. 
To construct such backdoor data, Turner~\etal~\cite{turner2019label-consistent} argue that the underlying image, before attaching the trigger, needs to become ``hard-to-classify.''
Intuitively, this choice would force the neural network to rely on the added trigger rather than the image semantics and hence, learn to associate the trigger with the target class more easily~\cite{turner2019label-consistent}.
To this end, Turner~\etal~\cite{turner2019label-consistent} first render hard-to-classify samples using adversarial perturbation or generative adversarial network interpolation and then add the trigger.
To further strengthen this attack, they reduce the trigger intensity (to go incognito) and install the trigger to all four corners of the poisoned image (to evade data augmentation).
Various stealthy trigger patterns can also help in constructing powerful clean-label attacks.
As such, different patterns like sinusoidal strips~\cite{barni2019sig}, invisible noise~\cite{zhong2020backdoor,li2020invisible}, natural reflections~\cite{liu2020refool}, and imperceptible warping~(WANet)~\cite{nguyen2021wanet} have been proposed as triggers.

In a different approach, Shafahi~\etal~\cite{shafahi2018frogs} use the idea of ``feature-collision'' to create clean-label poisoned data.
In particular, they try to find samples that are (1) similar to a base image and (2) close to the target class in the feature space of a pre-trained DNN.
This way, if the network is re-trained on the poisoned data, it will likely associate target class features to the poisoned data and hence, can fool the classifier.
Saha~\etal~\cite{saha2020htba} further extend ``feature-collision'' to data samples that are poisoned with patch triggers.
These triggers are installed at random locations in a given image.

%------------------------------------------------------------------------- 
\subsection{Backdoor Defense}
Existing backdoor defense techniques can be divided into several categories~\cite{goldlum2020dataset}.
Some methods aim at \textit{detecting backdoor poisoned data}~\cite{tran2018spectral,gao2019strip,chen2019detect,peri2020deepknn}.
Closely related to our work, Jin~\etal~\cite{jin2020aunified} try to detect backdoor samples using the local intrinsic dimensionality of the data.
To this end, they extract features of each image using a pre-trained neural network on clean data.
In contrast, as we shall see in \Cref{sec:proposed_method}, we use the feature space of the same neural network we are training, which may have been fed with poisoned data.
\textit{Identification of poisoned models} is another popular defense mechanism against backdoor attacks.
In this approach, given a DNN model, the aim is to detect if it has backdoors.
Neural cleanse~\cite{wang2019neuralcleanse}, DeepInspect~\cite{chen2019deepinspect}, TABOR~\cite{guo2019tabor}, and Universal Litmus Patterns~\cite{kolouri2020ulp} are some of the methods that fall into this category.
Furthermore, some techniques aim at \textit{removing the backdoor data effects in a trained neural network}~\cite{liu2018finepruning,wang2019neuralcleanse,liu2020removing,zhao2020bridging,yige2021nad}.

Most related to this work are approaches that try to \textit{avoid learning the triggers during training}~\cite{tran2018spectral,hayase2021defense}.
To this end, they first train a neural network on poisoned data.
Then, using the backdoored DNN and a clean validation set, they extract robust statistical features associated with clean samples.
Next, the training set is automatically inspected, and samples that do not meet the cleanness criteria are thrown away.
Finally, the neural network is re-trained on this new training dataset.
In contrast, our approach does not require additional certified clean data.
Moreover, our proposed method trains the neural network only once, taking less training time compared to existing methods~\cite{tran2018spectral,hayase2021defense}.
Other approaches in this category, such as deep partition aggregation~\cite{levine2021deep} and bagging~\cite{jia2020intrinsic}, require training multiple networks to enable a ``majority-vote mechanism''~\cite{goldlum2020dataset}.
However, our approach focuses on the robust training of a single neural network.

%===========================================================
\section{Background}\label{sec:background}
As mentioned in \Cref{sec:introduction}, our approach consists of a coreset selection algorithm based on the gradient space attributes and the local intrinsic dimensionality of the data.
This section reviews the background related to coreset selection and local intrinsic dimensionality.

%------------------------------------------------------------------------- 
\subsection{Coreset Selection}
Coreset selection refers to algorithms that create weighted subsets of the original data.
For deep learning, these subsets are selected so that training a model over them is approximately equivalent to fitting a model on the original data~\cite{mirzasoleiman2020craig}.
Closely related to this work, Mirzasoleiman~\etal~\cite{mirzasoleiman2020crust} exploit the idea of coreset selection for training with noisy labels.
It is argued that data with label noise would result in neural network gradients that differ from clean data.
As such, a coreset selection objective is defined to select the data with ``the most centrally located gradients''~\cite{mirzasoleiman2020crust}.
The network is then trained on this selected data.   

%------------------------------------------------------------------------- 
\subsection{Local Intrinsic Dimensionality}
Traditionally, classical expansion models such as generalized expansion dimension~(GED)~\cite{houle2012ged} were used to measure the intrinsic dimensionality of the data.
As a motivating example, consider two equicenter balls of radii $r_1$ and $r_2$ in a $d$-dimensional Euclidean space.
Assume the volumes of these balls are given as $V_1$ and $V_2$, respectively.
Then, the space dimension can be deduced using
\begin{equation}\label{eq:ged}\nonumber
    \frac{V_{2}}{V_{1}}=\left(\frac{r_{2}}{r_{1}}\right)^{d} \Rightarrow d=\frac{\ln \left(V_{2} / V_{1}\right)}{\ln \left(r_{2} / r_{1}\right)}.
\end{equation} 
To estimate the data dimension $d$, GED formulations approximate each ball's volume by the number of data samples they capture~\cite{karger2002finding,houle2012ged}.

By extending the aforementioned setting into a statistical one, classical expansion models can provide a local view of intrinsic dimensionality~\cite{houle2017LID_I,amsaleg2021high}.
To this end, the natural analogy of volumes and probability measures is exploited.
In particular, instead of a Euclidean space, a statistical setting powered with continuous distance distributions is considered.
\begin{definition}[\textbf{Local Intrinsic Dimensionality~(LID)}~\cite{houle2017LID_I,ma2018characterizing}]
	Let ${\boldsymbol{x} \in \mathbb{X}}$ be a data sample.
	Also, let $r>0$ denote a non-negative random variable that measures the distance of $\boldsymbol{x}$ to other data samples, and assume its cumulative distribution function is denoted by $F(r)$.
	If $F(r)$ is positive and continuously differentiable for every $r>0$, then the LID of $\boldsymbol{x}$ at distance $r$ is given by
	\begin{equation}\nonumber
	\operatorname{LID}_{F}(r) \triangleq \lim _{\epsilon \rightarrow 0^{+}} \frac{\ln (F((1+\epsilon) r) / F(r))}{\ln (1+\epsilon)}=\frac{r F^{\prime}(r)}{F(r)},
	\end{equation}
	whenever the limit exists.
	The LID at $\boldsymbol{x}$ is defined by taking the limit of $\operatorname{LID}_{F}(r)$ as $r \rightarrow 0^{+}$
	\begin{equation}\label{eq:lid_exact}
	\mathrm{LID}_{F}=\lim _{r \rightarrow 0^{+}} \operatorname{LID}_{F}(r).
	\end{equation}
\end{definition}
Calculating \Cref{eq:lid_exact} limit is not straightforward as it requires knowing the exact distance distribution $F(r)$.
Instead, several estimators using extreme value theory~(EVT) have been proposed~\cite{levina2004mlid,amsaleg2015estimating}.
Given its efficacy, here we use the following maximum likelihood estimator for LID~\cite{amsaleg2015estimating,ma2018characterizing}
\begin{equation}\label{eq:lid_estimate}
    \widehat{\operatorname{LID}}(\boldsymbol{x})=-\left(\frac{1}{k} \sum_{i=1}^{k} \log \frac{r_{i}(\boldsymbol{x})}{r_{k}(\boldsymbol{x})}\right)^{-1},
\end{equation}
where $r_{i}(\boldsymbol{x})$ is the distance between the data $\boldsymbol{x}$ and its $i$-th nearest neighbor from the dataset. 
As seen, only data samples in a local neighborhood would be considered in such an estimate.
Thus, under this regime, a \textit{local view} of the intrinsic dimensionality is obtained.

Finally, note that even computing the estimate given in \Cref{eq:lid_estimate} might become computationally prohibitive, especially for high-dimensional datasets with thousands of samples.
Thus, to further make this computation straightforward, Ma~\etal~\cite{ma2018d2l,ma2018characterizing} propose an alternative approach.
First, randomly sampled mini-batches of data are used to determine the set of nearest neighbors.
Second, instead of working with high-dimensional data samples directly, their feature space representation given by an underlying DNN is used.
We will also be using this approximator in our approach.

%===========================================================
\section{Proposed Method}\label{sec:proposed_method}

In this section, we formally define the problem of training neural networks with backdoor poisoned data. 
Next, we argue that two critical features can characterize clean samples in a backdoor poisoned dataset.
First, gradient updates for clean data differ in their magnitudes and directions from those of poisoned data~\cite{hong2020gradientshaping}.
Second, clean data usually has lower LID than its tampered counterparts~\cite{ma2018characterizing,jin2020aunified,amsaleg2021high}.
We define a COreset selection algorithm with LocaL Intrinsic DimEnisonality Regularization based on these two criteria, called \textsc{Collider}.
Using \textsc{Collider}, we effectively select data samples that satisfy the properties mentioned above and are likely to be clean.  
As such, the neural network trained on this subset of data would not be affected by the installed triggers.
As we shall see, \textsc{Collider} has several advantages over existing methods.
Namely, it does not require (1) having access to clean validation data, (2) re-training the whole model, or (3) training multiple neural networks.

%------------------------------------------------------------------------- 
\subsection{Problem Statement}\label{sec:sec:problem_statement}
Let $\mathcal{D}=\left\{\left(\boldsymbol{x}_{i}, y_{i}\right)\right\}_{i=1}^{n} \subset  \mathbb{X} \times \mathbb{C}$ denote a training dataset.
Each sample consists of an image $\boldsymbol{x}_{i}$ from domain $\mathbb{X}$ and a label $y_{i}$ which takes one of the $k$ possible values from the set $\mathbb{C}=\left[k\right]=\left\{1, 2, \dots, k\right\}$.
Suppose that ${f_{\boldsymbol{\theta}}: \mathbb{X} \rightarrow \mathbb{R}^{k}}$ denotes a neural network image classifier with parameters $\boldsymbol{\theta}$.
It takes an image $\boldsymbol{x}_{i} \in \mathbb{X}$ and outputs a $k$-dimensional real-valued vector, also known as the logit.
This vector gives the predicted label of the input $\boldsymbol{x}_{i}$ by $\hat{y}_{i}=\argmax f_{\boldsymbol{\theta}}\left(\boldsymbol{x}_{i}\right)$.
The neural network is trained to minimize an appropriately defined objective function
\begin{equation}\label{eq:nn_objective_loss}
    \mathcal{L}(\boldsymbol{\theta})=\sum_{i \in V}\ell\left(f_{\boldsymbol{\theta}}\left(\boldsymbol{x}_{i}\right), y_{i}\right)=\sum_{i \in V}\ell_{i}\left(\boldsymbol{\theta}\right),
\end{equation}
over the training set. Here, $\ell(\cdot)$ stands for a standard cost function such as cross-entropy, and $V=\left[n\right]=\left\{1, 2, \dots, n\right\}$ denotes the set of all training data.
Optimizing \Cref{eq:nn_objective_loss} using gradient descent requires finding the gradient of the loss for all the data points in $V$, which is costly.
In practice, one usually takes random mini-batches of training data and performs stochastic gradient descent.

Here, we assume that the training dataset consists of backdoor poisoned data.
In particular, let ${\mathcal{B}: \mathbb{X} \rightarrow \mathbb{X}}$ denote a backdoor data poisoning rule by which the data is tampered.\footnote{The adversary may also change the injected data labels. For notation brevity, we assume that the injection rule only changes the image itself, not the label.}
In targeted backdoor attacks, which we consider here, the adversary first replaces some training data samples $\boldsymbol{x}_{i}$ with their poisoned counterparts $\mathcal{B}\left(\boldsymbol{x}_{i}\right)$.
The goal is to force the model to unintentionally learn the injection rule, so that during inference for any test sample $\left(\boldsymbol{x}_{\text{test}}, y_{\text{test}}\right)$ the neural network outputs
\begin{align}\label{eq:backdoor_def}
    y_{\text{test}}=\argmax f_{\boldsymbol{\theta}}\left(\boldsymbol{x}_{\text{test}}\right)\quad \text{\&} \quad t=\argmax f_{\boldsymbol{\theta}}\left(\mathcal{B}\left(\boldsymbol{x}_{\text{test}}\right)\right),
\end{align}
where $t \in \mathbb{C}$ denotes the attacker's intended target class.
In this paper, we want to meticulously select a subset $S$ of the entire training data $V$ such that it only contains clean data samples.
To this end, we carefully identify properties of the clean data that should be in the set $S$.
We define a coreset selection objective based on these criteria to form the subset $S$ at each epoch.
The model is then trained on the data inside the coreset $S$.

%------------------------------------------------------------------------- 
\subsection{Clean vs. Poisoned Data Properties}\label{sec:sec:properties}
Next, we argue how the gradient space and LID differences between clean and poisoned samples can help us identify poisoned data.
Based on these arguments, we will then define a coreset selection objective to effectively filter out such samples from the training set in an online fashion.

%.........................................................................
\subsubsection{Gradient Space Properties.}\label{sec:sec:basic_coreset}

Recently, Hong~\etal~\cite{hong2020gradientshaping} have empirically shown that in the gradient space, poisoned data exhibit different behavior compared to clean samples.
Specifically, it is shown that the gradient updates computed on poisoned data have comparably (1) larger $\ell_2$ norm and (2) different orientation in contrast to clean data.
Given the empirical observations of~Hong~\etal~\cite{hong2020gradientshaping}, we conclude that the backdoor poisoned data have different gradients compared to the clean data. 
Moreover, Mirzasoleiman~\etal~\cite{mirzasoleiman2020crust} studied neural network training with noisy labeled data and argued that in the gradient space, clean data samples tend to be tied with each other.
Putting these two observations together, we deduce that the clean data samples are usually tied together in the gradient space while poisoned data is scattered in the gradient space. 
Thus, we can use this property to find the set of ``most centrally located samples'' in the gradient space and filter out poisoned data.

\begin{figure}[tb!]
	\centering
	\begin{subfigure}{.45\textwidth}
		\centering
		\captionsetup{justification=centering}
		\includegraphics[width=0.85\textwidth]{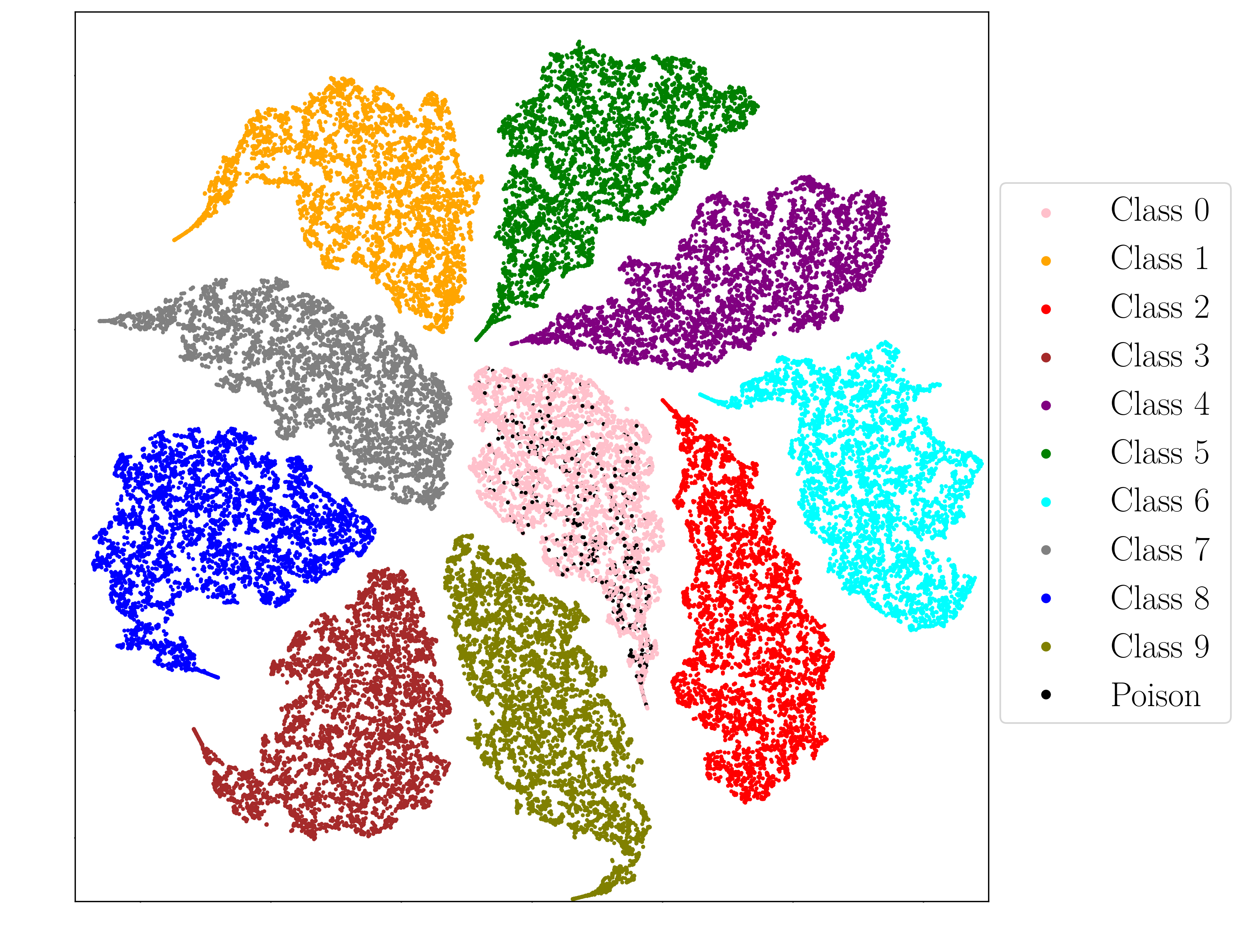}
		\caption{t-SNE~\cite{van2008visualizing} plot of a randomly initialized neural network gradient.}
		\label{fig:gradient:tsne}
	\end{subfigure}\hspace*{0.2em}
	\begin{subfigure}{.5\textwidth}
		\centering
		\captionsetup{justification=centering}
		\includegraphics[width=0.85\textwidth]{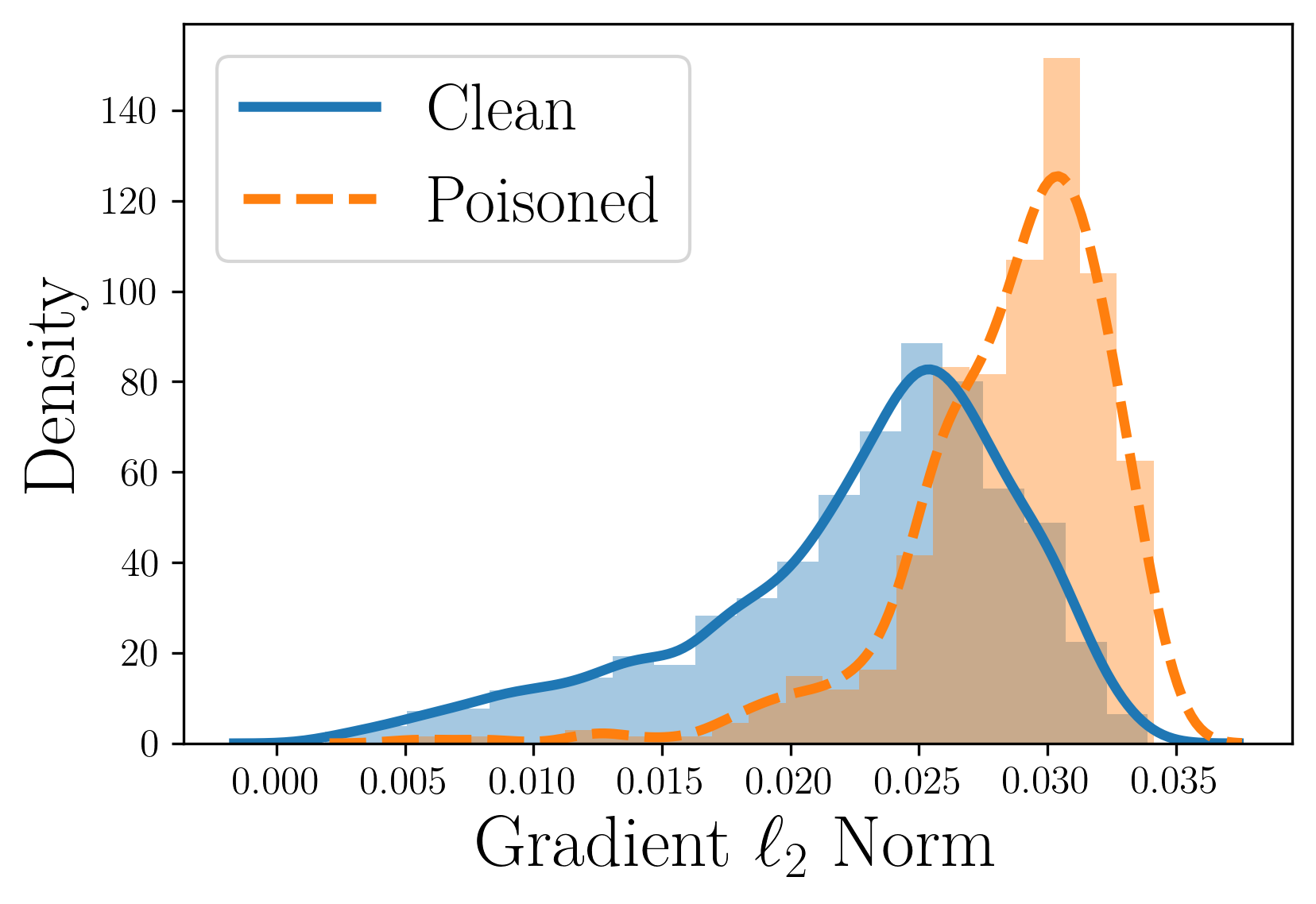}
		\caption{Distribution of the neural network gradient norm after 3 epochs of training.}
		\label{fig:gradient:dist}
	\end{subfigure}\hspace*{0.2em}
	 \caption{Visualization of the loss gradients for a neural network on poisoned CIFAR-10 dataset using the last layer approximation of \Cref{sec:sec:collider}. As seen, the gradients of the poisoned data are scattered in the gradient space and tend to exhibit a larger norm.}
	\label{fig:gradient}
	\vskip -0.20in
\end{figure}

We empirically investigate our assumptions around gradient space properties in \Cref{fig:gradient}. 
Specifically, we visualize the gradients of a randomly initialized neural network for the CIFAR-10 dataset poisoned with the backdoor triggers of~Gu~\etal~\cite{gu2017badnets}.
To this end, we first show the t-SNE~\cite{van2008visualizing} plot of the gradients in \Cref{fig:gradient:tsne}. 
As can be seen, the poisoned data is scattered in the gradient space.
Moreover, as seen in \Cref{fig:gradient:dist}, the $\ell_2$ norm of the gradients tends to be larger for the poisoned data.
This case study demonstrates how our assumptions around gradient space properties hold for poisoned data.

Unfortunately, as we will empirically see, after the first few epochs, the gradient dissimilarities between clean and poisoned samples gradually vanish.
This is in line with empirical observations of Hong~\etal~\cite{hong2020gradientshaping}.
As such, we need an additional regularizer to help us maintain the performance of data filtering throughout training.
To this end, we make use of LID.

%.........................................................................
\subsubsection{Local Intrinsic Dimensionality Properties.}\label{sec:sec:LID}
As discussed above, gradient space properties may not be sufficient to design a robust training algorithm against backdoor data.
Thus, we look into the geometrical structure of the underlying data to further enhance the performance of our method.
To this end, we utilize the LID of the data~\cite{houle2017LID_I,houle2017LID_II}.

Empirically, Ma~\etal~\cite{ma2018d2l} show that neural networks tend to progressively transform the data into subspaces of low local intrinsic dimensionality, such that each clean data point is surrounded by other clean samples.   
The LID has also been successfully applied to characterize adversarial subspaces~\cite{ma2018characterizing}.
In particular, Ma~\etal~\cite{ma2018characterizing} argue how the LID values of adversarially perturbed examples are probably higher than their underlying clean data.
This is motivated by the fact that legitimate perturbed examples are likely to be found in a neighborhood where not many clean examples reside, and hence, have high intrinsic dimensionality.
For $k$-NN classifiers, Amsaleg~\etal~\cite{amsaleg2021high} provide a rigorous theoretical analysis.
It has been proved that if data samples have large LID values, the level of perturbation required to turn them into adversarial examples diminishes.
This theoretical study supports the empirical observations of Ma~\etal~\cite{ma2018characterizing} that tampered examples exhibit high LID values.
These results collectively indicate that manipulating clean data samples will likely increase their LID.

\begin{figure}[tb!]
	\centering
	\begin{subfigure}{.45\textwidth}
		\centering
		\includegraphics[width=0.85\textwidth]{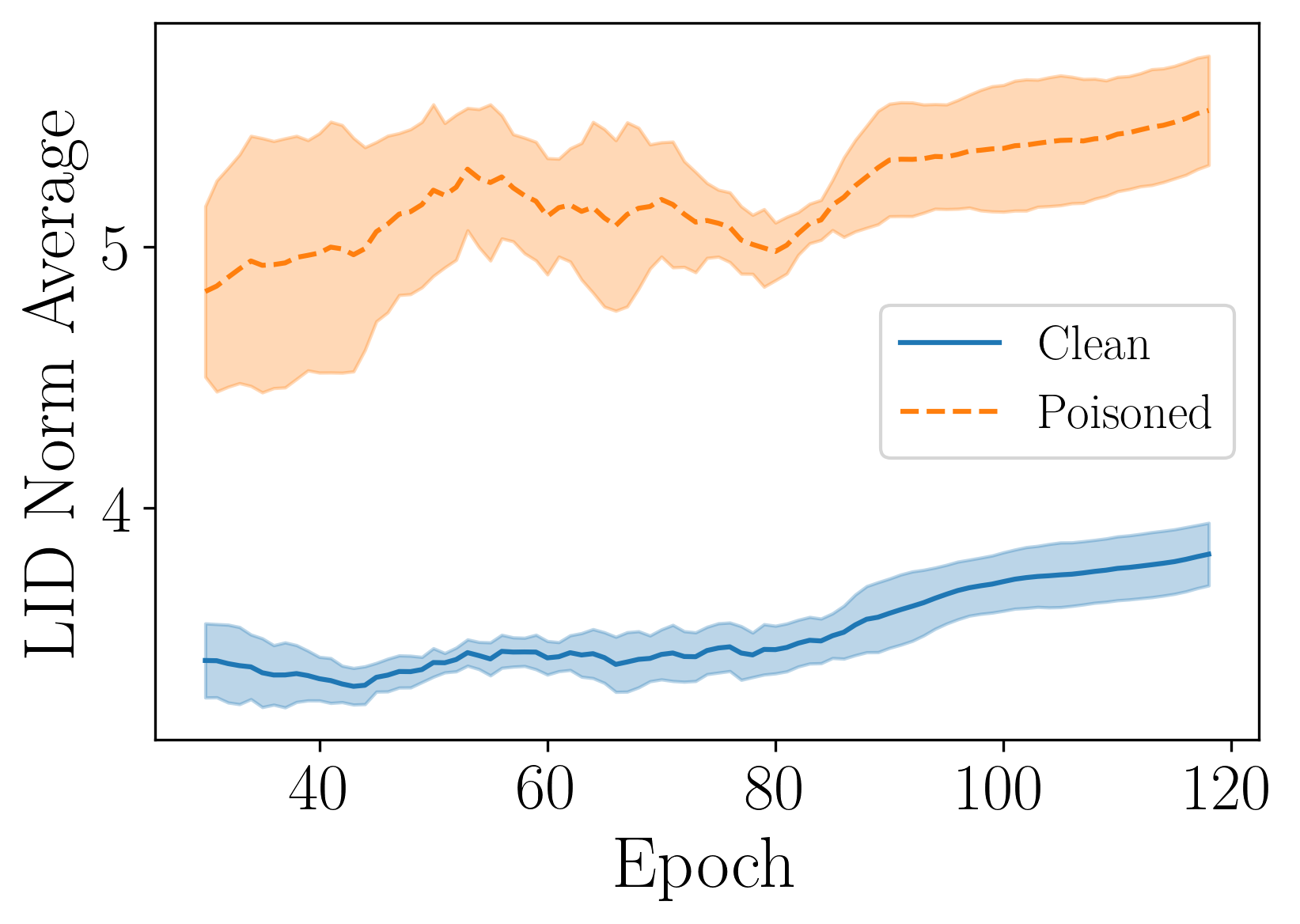}
		\caption{}
		\label{fig:LID:seeds}
	\end{subfigure}\hspace*{0.2em}
	\begin{subfigure}{.45\textwidth}
		\centering
		\includegraphics[width=0.85\textwidth]{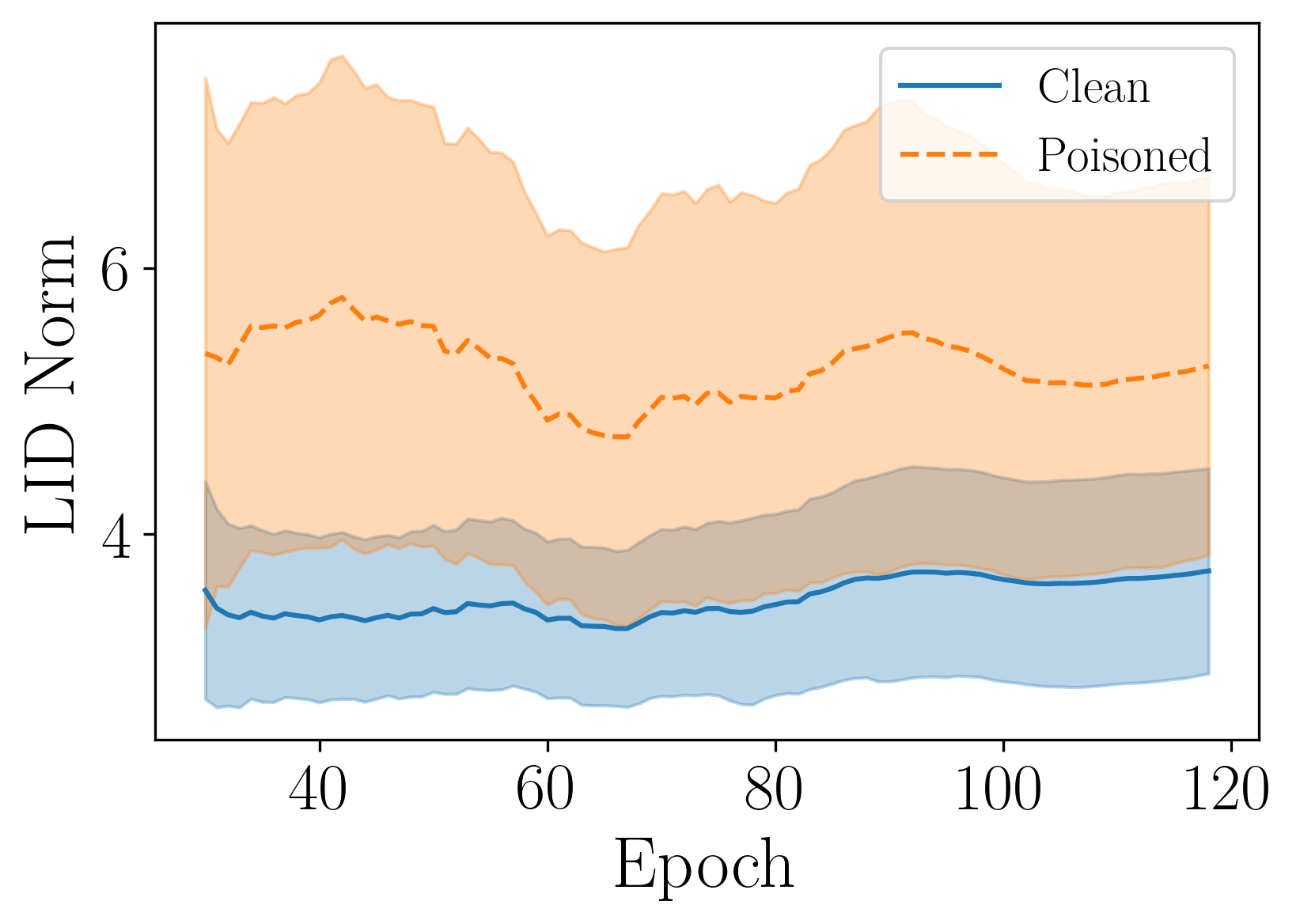}
		\caption{}
		\label{fig:LID:samples}
	\end{subfigure}\hspace*{0.2em}
	\caption{The LID of clean and BadNet poisoned data samples for CIFAR-10 dataset. (a) average LID norm across 5 different seeds (b) LID distribution for a single run. As expected, benign data samples have a lower LID compared to poisoned data.}
	\label{fig:LID}
	\vskip -0.20in
\end{figure}

Similarly, one would expect that successful backdoor attacks effectively move a sample away from all clean data.
This way, they try to eventually form a subspace secluded from the clean samples that can subvert the neural network's decision into the attacker's target class.
Thus, we contemplate that a neighborhood with higher dimensionality is needed to shelter poisoned samples compared to clean data.
In other words, clean data samples are expected to have low LID values.
In \Cref{fig:LID}, we empirically show the distribution of LID values for clean and poisoned data recorded during training a neural network on the CIFAR-10 dataset.
As expected, poisoned data generally exhibit a higher LID.

%------------------------------------------------------------------------- 
\subsection{\textsc{COLLIDER}: COreset selection with LocaL Intrinsic DimEnisonality Regularization}\label{sec:sec:collider}

In this section, we exploit the properties of clean data discussed in \Cref{sec:sec:properties} and define our coreset selection objective.
As stated in \Cref{sec:sec:properties}, we need to first identify the set of most centrally located samples in the gradient space as the clean data are likely to be tied together.
Thus, as our base component, we use a coreset selection objective to find clusters of data that are tied together in the gradient space~\cite{mirzasoleiman2020crust}
\begin{equation}\label{eq:basic_coreset}
S^{*}(\boldsymbol{\theta}) \in \argmin_{S \subseteq V} \sum_{i \in V} \min _{j \in S} d_{i j}(\boldsymbol{\theta}) \quad \text { s.t. } \quad |S| \leq k.
\end{equation}
Here, $d_{i j}(\boldsymbol{\theta})=\left\|\nabla\ell_{i}\left(\boldsymbol{\theta}\right)-\nabla \ell_{j}\left(\boldsymbol{\theta}\right)\right\|_{2}$ shows the $\ell_{2}$ distance of loss gradients between samples $i$ and $j$, and $k$ denotes the number of samples in the coreset.

To add the LID to our approach, we need to ensure that the underlying neural network has reached a point where it can extract meaningful features from the data.
Remember from \Cref{sec:background} that this is important as these features are used to approximate the LID values, and thus, they need to have a meaningful structure across data.
Therefore, we run our basic coreset selection objective of \Cref{eq:basic_coreset} during the first epochs until we obtain a steady validation accuracy.
Afterward, we start to consider the LID values.
Remember that we want to encourage the selection of samples that are likely to be clean, i.e., have low LID values.
Thus, we add a regularizer to our coreset selection objective that promotes this.
In particular, \Cref{eq:basic_coreset} is re-written as
\begin{equation}\label{eq:collider}
    S^{*}(\boldsymbol{\theta}) \in \argmin_{S \subseteq V, |S| \leq k} \sum_{i \in V} \min _{j \in S} d_{i j}(\boldsymbol{\theta}) + \lambda \mathrm{LID}\left(\boldsymbol{x}_{j}\right),
\end{equation}
where $\lambda$ is a hyperparameter that determines the relative importance of LID against the gradient term.
Here, $\mathrm{LID}\left(\cdot\right)$ is computed as discussed in \Cref{sec:background} using mini-batches of data.
As seen in \Cref{eq:collider}, we encourage the algorithm to select data points with low LID values.  

%------------------------------------------------------------------------- 
\subsection{Practical Considerations}

\paragraph{Gradient Approximation and Greedy Solvers.}
As pointed out in~\cite{mirzasoleiman2020craig,mirzasoleiman2020crust}, finding the optimal solution of \Cref{eq:basic_coreset} and as a result \Cref{eq:collider} is intractable.
This is since computing $d_{i j}(\boldsymbol{\theta})$ for all $i \in V$ requires backpropagation over the entire training set, which is computationally prohibitive.
More importantly, finding the optimal coresets as in \Cref{eq:basic_coreset} is shown to be NP-hard~\cite{mirzasoleiman2020craig}.

To address the first issue, the following upper-bound is usually used instead of the exact $d_{i j}(\boldsymbol{\theta})$ values~\cite{katharopoulos2018notall,mirzasoleiman2020craig,mirzasoleiman2020crust}
\begin{footnotesize}
	\begin{align}\label{eq:upper_bound}
	d_{i j}(\boldsymbol{\theta})&=\left\|\nabla\ell_{i}\left(\boldsymbol{\theta}\right)-\nabla \ell_{j}\left(\boldsymbol{\theta}\right)\right\|_{2} \nonumber\\
	&\leq c_{1}\left\|\Sigma_{L}^{\prime}\left(z_{i}^{(L)}\right) \nabla \ell_{i}^{(L)}(\boldsymbol{\theta})-\Sigma_{L}^{\prime}\left(z_{j}^{(L)}\right) \nabla \ell_{j}^{(L)}(\boldsymbol{\theta})\right\|+c_{2},
	\end{align}
\end{footnotesize}where $\Sigma_{L}^{\prime}\left(z_{i}^{(L)}\right) \nabla \ell_{i}^{(L)}(\boldsymbol{\theta})$ denotes the gradient of the loss with respect to the neural network's penultimate layer, shown by $z_{i}^{(L)}$.
Also, $c_1$ and $c_2$ are constants, and $L$ denotes the number of DNN layers.
This upper-bound can be computed efficiently, with a cost approximately equal to forward-propagation~\cite{mirzasoleiman2020crust}.
As for the NP-hardness issue, there exist efficient greedy algorithms that can solve \Cref{eq:basic_coreset} sub-optimally~\cite{minoux1978accelerated,nemhauser1978analysis,wolsey1982greedy}.
Specifically, \Cref{eq:basic_coreset} is first turned into its \textit{submodular optimization} equivalent, and then solved.
Details of this re-formulation can be found in~\cite{mirzasoleiman2020craig,mirzasoleiman2020crust} and \Cref{sec:ap:collider}.

\paragraph{Repetitive Coreset Selection.}
As our coresets are a function of the neural network parameters $\boldsymbol{\theta}$, we need to update them as the training goes on.
Thus, at the beginning of every epoch, we first select data samples $S$ by solving \Cref{eq:collider}.
We then train the neural network using mini-batch gradient descent on the data that falls inside the coreset $S$.

\paragraph{LID Computation.}
For LID estimation, we select neighbors of each sample from the same class.
This way, we ensure that clean samples have similar feature representations among themselves.
As a result, our LID estimates can better capture the dimensionality differences between the clean and poisoned samples.
To further stabilize LID estimates, we use a moving average of LID values taken over a few successive epochs, not a single one, as the LID estimate of a training sample. 
Moreover, as the training evolves, we permanently eliminate samples with the highest LID values.
This is motivated by the fact that our LID estimates are the average of LID values for the past epochs.
Thus, samples that exhibit a high LID value over successive training epochs are likely to be poisoned data.
However, we must be careful not to remove too many samples so that the solution to \Cref{eq:collider} becomes trivial.
As a result, we select the number of removals such that the coreset selection always has a non-trivial problem to solve.
According to \Cref{eq:collider}, this means that at the last epoch, we need to have at least $k$ samples left in our training set.
Thus, we gradually eliminate $n-k$ samples throughout the training.
Also, as in Mirzasoleiman~\etal~\cite{mirzasoleiman2020crust}, we use mix-up to enhance the quality of our coresets further.
Finally, note that the computational complexity of the LID regularization is minimal.
This is because for computing the $d_{i j}(\boldsymbol{\theta})$ terms in \Cref{eq:basic_coreset} we have already computed the logits.
The same values can be re-used for LID calculation.
The final algorithm can be found in \Cref{sec:ap:collider}.

%===========================================================
\section{Experimental Results}\label{sec:experiments}

This section presents our experimental results.
We demonstrate how by using \textsc{Collider} one can train robust neural networks against backdoor data poisonings.
Furthermore, we show the role of each component in \textsc{Collider} in our extensive ablation studies.

%-------------------------------------------------------------------------
\subsubsection{Settings.}
We use CIFAR-10~\cite{krizhevsky2009learning}, SVHN~\cite{netzer2011reading}, and ImageNet-12~\cite{russakovsky2015imagenet,liu2020refool} datasets in our experiments.
To test our approach against various backdoor data poisonings, we use BadNets~\cite{gu2017badnets}, label-consistent attacks~\cite{turner2019label-consistent}, sinusoidal strips~\cite{barni2019sig}, and triggers used in HTBA~\cite{saha2020htba} to poison the aforementioned datasets.
We randomly select a target class for each dataset and poison it with backdoor-generated data.
The ratio of poisoned samples in the target class is referred to as the \textit{injection rate}.
Samples from each backdoor poisoned dataset can be found in \Cref{sec:ap:imp_det}.
Furthermore, we use ResNet~\cite{he2016deep} as our DNN architecture.
For training, we use a stochastic gradient descent~(SGD) optimizer with momentum of 0.9 and weight decay~${5\text{e-}4}$.
The initial learning rate is set to 0.1, which is divided by 10 at epochs 80 and 100 for CIFAR-10 and SVHN~(epochs 72 and 144 for ImageNet-12).
In what follows, the \textit{coreset size} refers to the ratio of the original data in each class that we intend to keep for training.
For more details on the experimental settings and also extra experiments see \Cref{sec:ap:imp_det,sec:ap:ext_sim_res}.

%-------------------------------------------------------------------------
\subsubsection{Baselines.}
Our approach differs significantly in its assumptions from the current backdoor defenses.
Thus, for a fair comparison, we compare \textsc{Collider} to two other baselines: (1) the usual training, denoted as ``vanilla'', and (2) basic coreset selection~(\Cref{eq:basic_coreset}), denoted as ``coresets''.
Furthermore, to provide a complete picture of the performance of our proposed approach, we also include Spectral Signatures~(SS)~\cite{tran2018spectral}, Activation Clustering~(AC)~\cite{chen2019detect}, \textsc{Spectre}~\cite{hayase2021defense}, and Neural Attention Distillation~(NAD)~\cite{yige2021nad} to our comparisons.
It should be noted that these approaches have a different set of assumptions compared to \textsc{Collider}, and the comparisons are not fair.
This is because SS~\cite{tran2018spectral}, AC~\cite{chen2019detect}, and \textsc{Spectre}~\cite{hayase2021defense} train the neural network \textit{twice}, and NAD~\cite{yige2021nad} is trying to erase the backdoor after the training is complete.

\begin{table*}[tb!]\setlength{\tabcolsep}{4.5pt}
	\caption{Clean test accuracy (ACC) and attack success rate (ASR) in \% for backdoor data poisonings on CIFAR-10~(BadNets and label-consistent) and SVHN (sinusoidal strips) datasets.
		The results show the mean and standard deviation for 5 different seeds.
		The poisoned data injection rate is 10\%.
		For BadNets and label-consistent attacks, the coreset size is 0.3. It is 0.4 for sinusoidal strips.}
	\label{tab:small_results}
	\begin{center}
		\begin{tiny}
			\begin{tabular}{ccccccc}
				\toprule
				\multirow{2}{*}{Training}    & \multicolumn{2}{c}{BadNets~\cite{gu2017badnets}}  & \multicolumn{2}{c}{Label-consistent~\cite{turner2019label-consistent}}  & \multicolumn{2}{c}{Sinusoidal Strips~\cite{barni2019sig}} \\
				\cmidrule(lr){2-3} \cmidrule(lr){4-5} \cmidrule(lr){6-7}
				& ACC               & ASR               & ACC               & ASR             & ACC               & ASR\\
				\midrule
				Vanilla                                                                 & $92.19 \pm 0.20$	& $99.98 \pm 0.02$	& $92.46 \pm 0.16$	& $100$             & $95.79 \pm 0.20$  & $77.35 \pm 3.68$\\
				\midrule
				SS~\cite{tran2018spectral}                                              & $92.05 \pm 0.43$	& $1.18 \pm 0.36$	& $92.24 \pm 0.35$	& $0.53 \pm 0.09$   & $95.38 \pm 0.28$  & $77.30 \pm 2.50$\\
				AC~\cite{chen2019detect}                                                & $91.78 \pm 0.21$  & $99.87 \pm 0.08$  & $91.59 \pm 0.31$  & $75.44 \pm 42.53$ & $95.45 \pm 0.20$  & $77.43 \pm 4.59$\\ 
				\textsc{Spectre}~\cite{hayase2021defense}                               & $91.28 \pm 0.22$	& $98.17 \pm 1.97$  & $91.78 \pm 0.37$	& $0.51 \pm 0.15$   & $95.41 \pm 0.12$  & $8.51 \pm 7.03$\\
				\midrule
				NAD~\cite{yige2021nad}                                                  & $72.19 \pm 1.73$	& $3.55 \pm 1.25$	& $70.18 \pm 1.70$	& $3.44 \pm 1.50$ & $92.41 \pm 0.34$  & $6.99 \pm 3.02$\\
				\midrule
				Coresets                                                                & $84.86 \pm 0.47$  & $74.93 \pm 34.6$  & $83.87 \pm 0.36$  & $7.78 \pm 9.64$ & $92.30 \pm 0.19$  & $24.30 \pm 8.15$\\ 
				\textsc{Collider} (Ours)                                                & $80.66 \pm 0.95$	& $4.80 \pm 1.49$   & $82.11 \pm 0.62$	& $5.19 \pm 1.08$ & $89.74 \pm 0.31$  & $6.20 \pm 3.69$\\
				\bottomrule
			\end{tabular}
		\end{tiny}
	\end{center}
	\vskip -0.30in
\end{table*}

%-------------------------------------------------------------------------
\subsection{Clean Test Accuracy and Attack Success Rate}
In \Cref{tab:small_results} we show our experimental results.
We measure the performance of each training algorithm by two quantities: (1) accuracy on the clean test set~(ACC) and (2) the attack success rate~(ASR), which indicates the accuracy of the installed triggers on non-target class images.
As seen, using \textsc{Collider} we can reduce the attack success rate significantly in nearly all the cases.
Moreover, although the coreset selection based on the gradient space properties can improve robustness (see the ``coresets'' row), its performance is insufficient to provide a robust model.
This, as we see, can be improved using the intrinsic dimensionality of the data that captures the geometrical structure of the training samples.

Interestingly, observe how our approach is a middle-ground between SS~\cite{tran2018spectral} and NAD~\cite{yige2021nad} settings.
Specifically, in SS~\cite{tran2018spectral} the entire network is re-trained to get a better performance, but this is inefficient.
To show this, in \Cref{tab:small_results_time} we report the total training time for our approach against SS~\cite{tran2018spectral} that trains the neural network twice.
As seen, our approach results in an average of $22$\% reduction in training time compared to existing methods that need re-training.\footnote{Note that here we measure the training time for SS~\cite{tran2018spectral} as a representative of techniques that need re-training, e.g., AC~\cite{chen2019detect} and \textsc{Spectre}~\cite{hayase2021defense}.}

At the other end of the spectrum, NAD~\cite{yige2021nad} tries to erase the backdoors from the vanilla network without re-training.
In \textsc{Collider} we are trying to prevent the network from learning the triggers in an online fashion, such that there would be no need to re-train the network.
More importantly, NAD~\cite{yige2021nad} shows the difficulty of this task because once the model learns the triggers, one cannot erase them without sacrificing too much of the clean accuracy.

As seen in \Cref{tab:small_results}, while \textsc{Collider} can significantly reduce the ASR, it also exhibits a gap in terms of ACC with other methods. As a remedy, we investigate using non-coreset data in training to increase the clean accuracy gap while maintaining backdoor robustness.
To this end, we use semi-supervised learning and treat the non-coreset data as unlabeled samples.
As shown in \Cref{sec:ap:ext_sim_res}, one can enhance the ACC while decreasing ASR.
For a detailed description of this version of our approach and other extensive experiments, please see \Cref{sec:ap:ext_sim_res}.

\begin{table*}[tb!]\setlength{\tabcolsep}{4.5pt}
	\caption{Total training time (in minutes) for experiments of \Cref{tab:small_results}.
		The results show the mean and standard deviation for 5 different seeds.}
	\label{tab:small_results_time}
	\begin{center}
		\begin{scriptsize}
			\begin{tabular}{cccc}
				\toprule
				Method                                        & BadNets~\cite{gu2017badnets} & Label-consistent~\cite{turner2019label-consistent}  & Sinusoidal Strips~\cite{barni2019sig}\\
				\midrule
				Coresets                                      & $61.35 \pm 0.31$             & $63.09 \pm 0.36$                                    & $66.04 \pm 1.11$\\ 
				\textsc{Collider}                             & $62.56 \pm 0.13$             & $67.10 \pm 0.95$                                    & $64.53 \pm 0.38$\\
				\textsc{SS}~\cite{tran2018spectral}~($\sim$AC~\cite{chen2019detect}/\textsc{Spectre}~\cite{hayase2021defense})          & $85.48 \pm 0.28$             & $85.26 \pm 0.26$                                    & $79.46 \pm 0.86$\\
				\bottomrule
			\end{tabular}
		\end{scriptsize}
	\end{center}
	\vskip -0.20in
\end{table*}

\begin{figure*}[tb!]
	\centering
	\begin{subfigure}{.30\textwidth}
		\centering
		\includegraphics[width=1.0\textwidth]{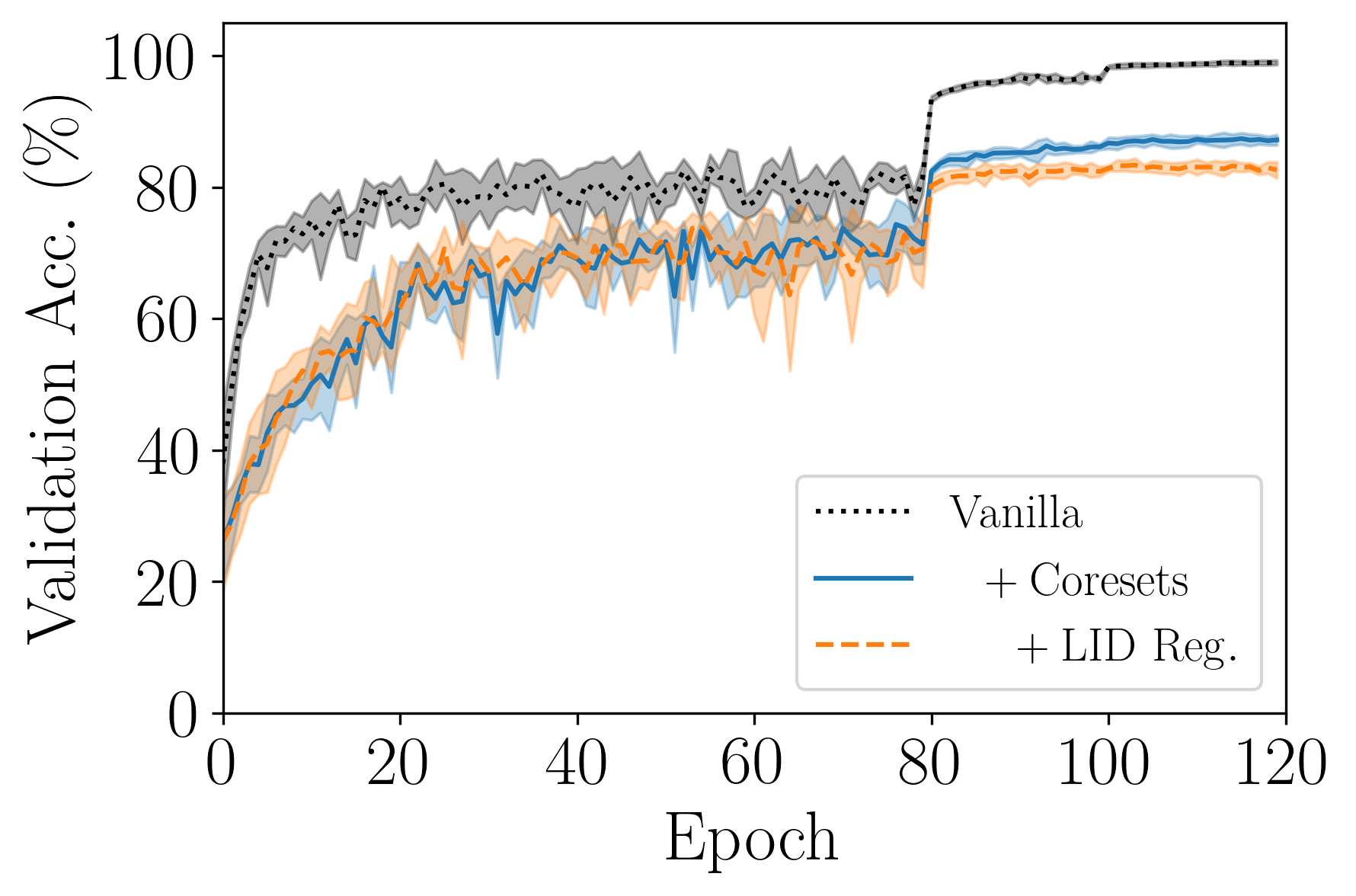}
		\caption{}
		\label{fig:ablation:val_acc}
	\end{subfigure}\hspace*{0.2em}
	\begin{subfigure}{.30\textwidth}
		\centering
		\includegraphics[width=1.0\textwidth]{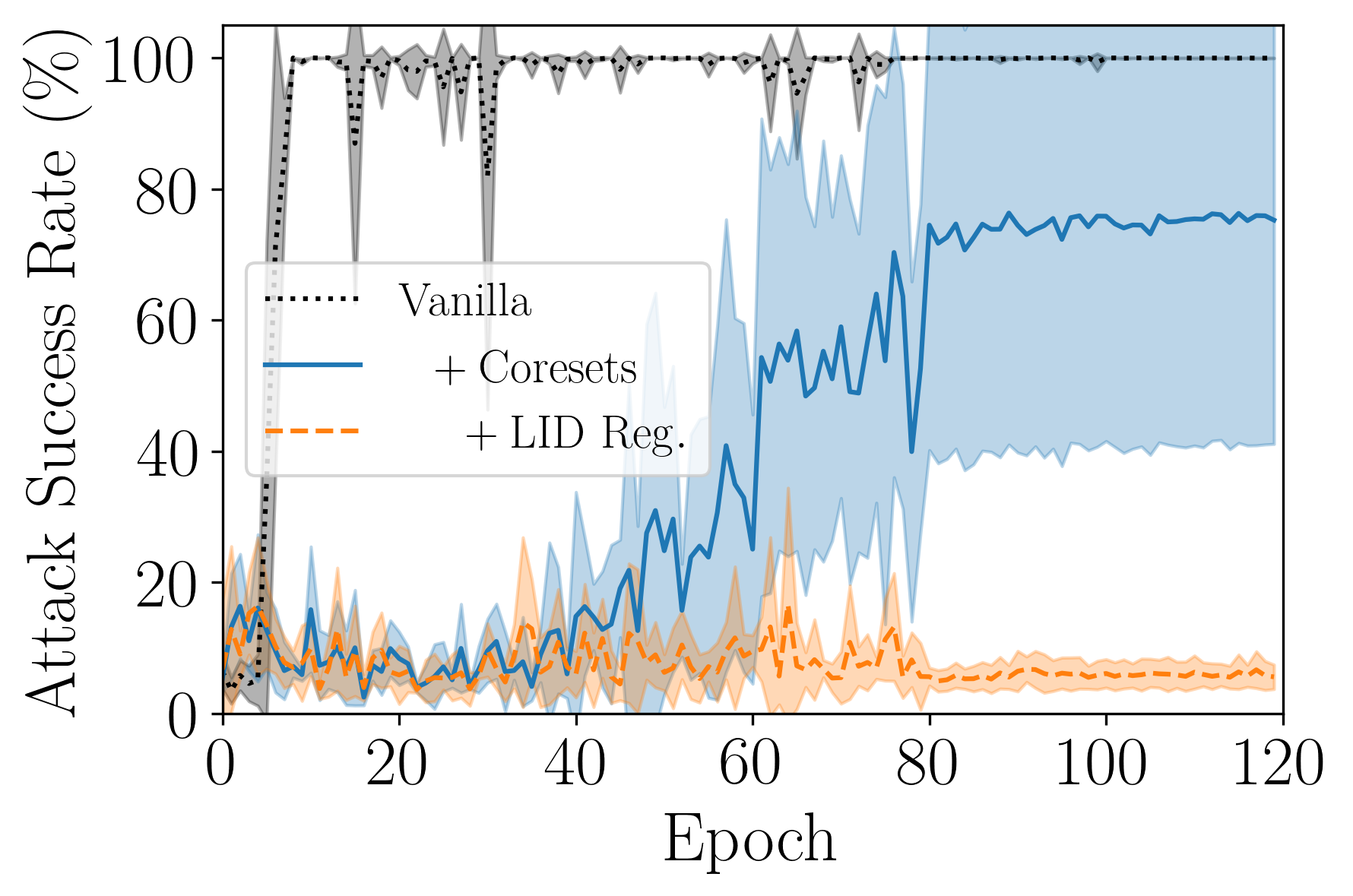}
		\caption{}
		\label{fig:ablation:asr}
	\end{subfigure}\hspace*{0.2em}
	\begin{subfigure}{.30\textwidth}
		\centering
		\includegraphics[width=1.0\textwidth]{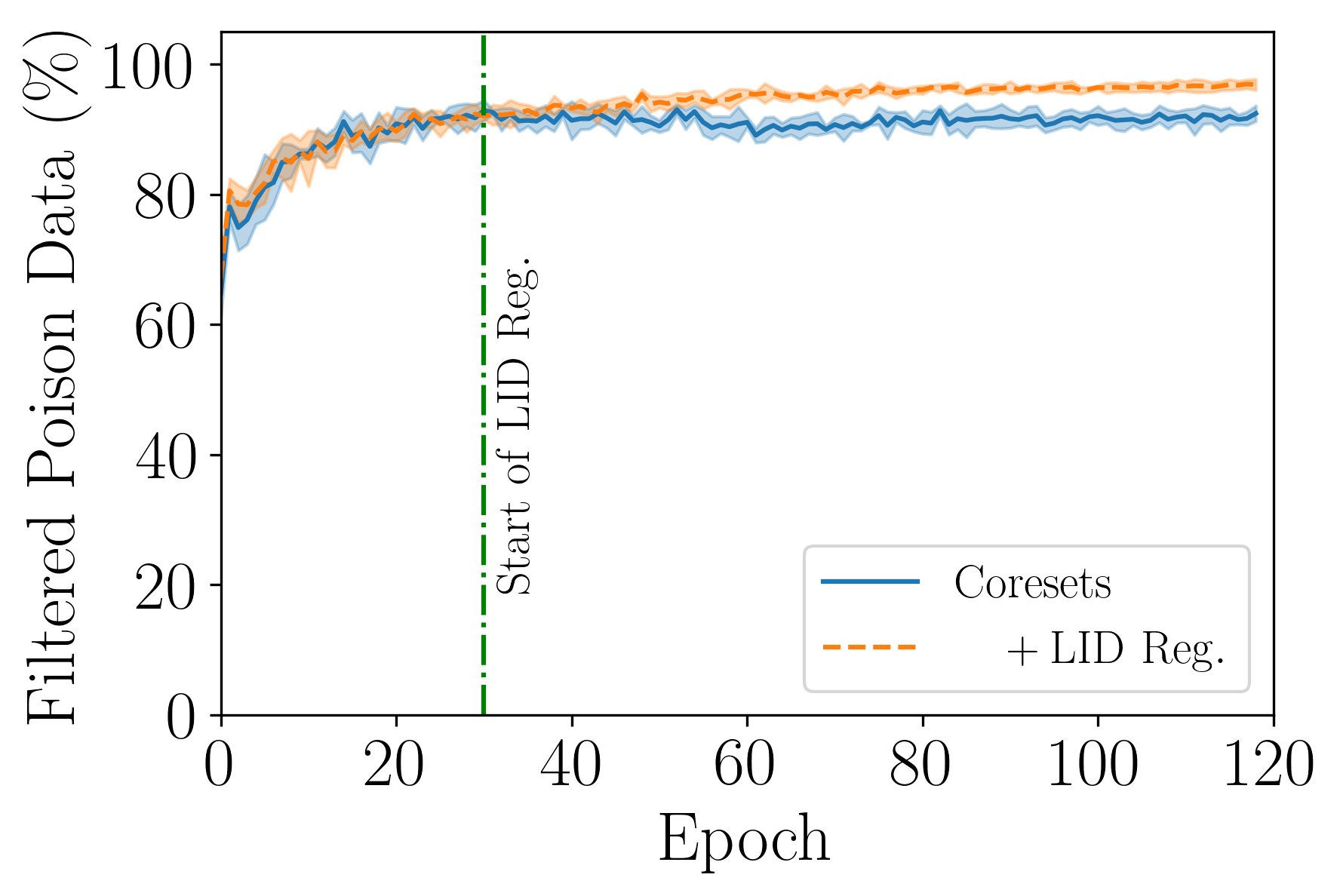}
		\caption{}
		\label{fig:ablation:label_acc}
	\end{subfigure}
	\vskip -0.1 in
	\caption{Ablation study on the effect of coreset selection and LID regularization terms in \textsc{Collider} for training robust models on BadNet poisonings. (a) validation accuracy (b) attack success rate (c) percentage of poisoned data filtered. Epoch 30 (annotated in (c)) is the start of our LID regularization. As seen, while the difference between the percentage of the filtered poison data is very close, this translates to a significant difference in ASR.}
	\label{fig:ablation}
	\vskip -0.135in
\end{figure*}

%-------------------------------------------------------------------------
\subsection{Ablation Studies}
To clarify the role of each component in \textsc{Collider}, we record some performance measures during training.
For each algorithm, we record the accuracy on the validation set and the attack success rate on the test set.
Also, to better monitor the coresets, we record the percentage of the poisoned data filtered during training.
This measure is defined as the complement of the number of poisoned data in the coreset divided by the total number of poisoned data.
This ratio is always between zero and one: one means that the selected coreset does not contain any poisoned data.
As seen in~\Cref{fig:ablation:label_acc}, after the first few epochs where the performance of the basic coreset selection improves, it suddenly degrades.
At this point, we are ready to enable the LID regularization in \textsc{Collider} since we have reached a point with steady validation accuracy.
This addition helps the coreset selection to maintain its performance throughout the training.
This slight difference is decisive when it comes to the attack success rate, which, as seen in \Cref{fig:ablation:asr}, can determine the robustness of the final model.\footnote{Note that the sudden jump observed in the validation accuracy during epochs 80 and 100 is due to the use of a multi-step learning rate scheduler discussed at the beginning of \Cref{sec:experiments}.}

%-------------------------------------------------------------------------
\subsection{Trade-offs}
Finally, there are some trade-offs in \textsc{Collider} that need to be discussed.
First, in \Cref{fig:ap:coreset_size_trade-off} in \Cref{sec:ap:ext_sim_res} one can see that as we increase the coreset size, the clean test accuracy and the attack success rate both increase.
This is inevitable as by increasing the coreset size, we are likely to let more poisoned data into the selected subset used for training.
As a result, the attack success rate starts to increase.
However, we can see that this trade-off is less severe in \textsc{Collider} due to its use of LID that can capture the poisoned samples more accurately.
Second, in \Cref{fig:ap:asr_vs_inj} of \Cref{sec:ap:ext_sim_res} we show the performance of our proposed method as the injection rate is varied.
For the experiments in this figure, the coreset size is fixed to 0.3 or 30\%.
Ideally, we should be able to train a robust model with a 70\% injection rate successfully.
However, as discussed in \Cref{sec:sec:basic_coreset}, our solutions to the coreset selection objective of \Cref{eq:collider} are always sub-optimal as finding the exact solution is NP-hard.
Thus, as the injection rate increases, so does the attack success rate.
Again, the rate of this trade-off is less dramatic for \textsc{Collider} compared to gradient-based coreset selection.

%===========================================================
\section{Conclusion}\label{sec:conclusion}

We proposed a robust, end-to-end training algorithm for backdoor datasets in this paper.
Our method, named \textsc{Collider}, exploits the idea of coreset selection to purify the given data before using them for training a neural network.
Specifically, we saw how clean and poisoned data samples differ in their gradient space and local intrinsic dimensionality attributes.
Based on these differences, we defined a coreset selection algorithm that tries to filter out malicious data effectively.
The neural network can then be trained on this carefully selected data.
We showed the performance of the proposed approach in improving the robustness of neural networks against backdoor data poisonings and identified the role of each component through extensive ablation studies.
While successful in decreasing the attack success rate significantly, we saw that \textsc{Collider} also reduces the clean test accuracy slightly.

\noindent \subsubsection{Acknowledgments.} We thank Michael Houle for thoughtful discussions and comments on LID evaluation.
This research was undertaken using the LIEF HPC-GPG\-PU Facility hosted at the University of Melbourne.
This Facility was established with the assistance of LIEF Grant LE170100200.
Sarah Erfani is in part supported by Australian Research Council~(ARC) Discovery Early Career Researcher Award~(DECRA) DE220100680.

\bibliographystyle{splncs}
\bibliography{references}

%===========================================================
\newpage
\appendix
\section{\textsc{COLLIDER} Details}\label{sec:ap:collider}
In this section, first we show how the coreset selection objective of \Cref{eq:collider} can be turned into a submodular maximization equivalent.
This re-formulation enables us to use greedy algorithms to solve the coreset selection objective.
Afterwards, we present the full \textsc{Collider} algorithm in detail.

%-------------------------------------------------------------------------
\subsection{The Submodular Maximization Equivalent of the Coreset Selection Objective~\cite{mirzasoleiman2020craig,mirzasoleiman2020crust}}\label{sec:ap:sec:submodular}
Without loss of generality, we use the derivation of Mirzasoleiman~\etal~\cite{mirzasoleiman2020crust} for the basic coreset selection objective of \Cref{eq:basic_coreset}.
The equivalency of \Cref{eq:collider} to submodular maximization can be deduced similarly.

A set function $F: 2^{\left|V\right|} \rightarrow \mathbb{R}^{+}$ is called \textit{submodular} if for any $S \subset T \subset V$ and $e \in V \backslash T$ we have~\cite{mirzasoleiman2020crust}
\begin{equation}\nonumber
F\left(S \cup \left\{e\right\}\right) - F\left(S\right) \geq F\left(T \cup \left\{e\right\}\right) - F\left(T\right).
\end{equation}
Moreover, if for any $S \subset V$ and $e \in V \backslash S$ we have a non-negative $F\left(e|S\right)$, then $F(\cdot)$ is called \textit{monotone}~\cite{mirzasoleiman2020crust}.

Now, assume that we can find a \textit{constant upper-bound} for all the $d_{i j}(\boldsymbol{\theta})$ values from \Cref{eq:basic_coreset}.
If we denote this constant upper-bound with $d_{0}$, then \Cref{eq:basic_coreset} can be re-written as~\cite{mirzasoleiman2020crust}
\begin{equation}\label{eq:submodular_equi}
S^{*}(\boldsymbol{\theta}) \in \arg \max _{S \subseteq V} \sum_{i \in V} \max _{j \in S} d_{0} - d_{i j}^{u}(\boldsymbol{\theta}) \quad \text { s.t. } \quad |S| \leq k,
\end{equation}
where
$d_{i j}^{u}(\boldsymbol{\theta})=\left\|\Sigma_{L}^{\prime}\left(z_{i}^{(L)}\right) \nabla \ell_{i}^{(L)}(\boldsymbol{\theta})-\Sigma_{L}^{\prime}\left(z_{j}^{(L)}\right) \nabla \ell_{j}^{(L)}(\boldsymbol{\theta})\right\|$
is the \textit{$\boldsymbol{\theta}$-dependent upper-bound} of $d_{i j}(\boldsymbol{\theta})$ from \Cref{eq:upper_bound}.
Mirzasoleiman~\etal~\cite{mirzasoleiman2020crust} argue that the function
\begin{equation}\label{eq:submodular_func}\nonumber
F(S, \boldsymbol{\theta}) = \sum_{i \in V} \max _{j \in S} d_{0} - d_{i j}^{u}(\boldsymbol{\theta})
\end{equation}
is a \textit{monotone submodular} set function.
As such, \Cref{eq:submodular_equi} is equivalent to a well-known submodular maximization problem widely known as the \textit{facility location}, and there exist efficient greedy algorithms that can solve it sub-optimally.\footnote{In our experiments, we use the implementation of~Mirzasoleiman~\etal~\cite{mirzasoleiman2020crust} available \href{https://github.com/snap-stanford/crust}{online}.} 

%-------------------------------------------------------------------------
\subsection{COLLIDER Final Algorithm}
\Cref{alg:collider} shows the \textsc{Collider} framework for training neural networks on backdoor poisoned data in detail.
\begin{algorithm}[htp]
	\vspace*{0.25em}
	\caption{\textsc{Collider} for training DNNs on backdoor poisoned data\label{alg:collider}} 
	\textbf{Input}: dataset $\mathcal{D}=\left\{\left(\boldsymbol{x}_{i}, y_{i}\right)\right\}_{i=1}^{n}$, neural network~${f_{\boldsymbol{\theta}}(\cdot)}$.\vspace*{0.25em}
	\\
	\textbf{Output}: robustly trained neural network~${f_{\boldsymbol{\theta}}(\cdot)}$.\vspace*{0.25em}
	\\
	\textbf{Parameters}: number of classes~$C$, total epochs~$T$, batch size~$b$, coreset size~$k$, LID start epoch~$l$, LID number of neighbors~$N$, LID moving average window $w$, LID Lagrange multiplier $\lambda$.\vspace*{0.25em}
	\begin{algorithmic}[1]
		\State Initialize~${\boldsymbol{\theta}}$ randomly.
		\For {$t=1,2,\ldots, T$}
		\State $S^{t}=\emptyset$
		\For {$c=1,2,\ldots, C$}\vspace*{0.15em}
		\State $\mathcal{D}^{c} = \cup_{i=1}^{n}\left\{\left(\boldsymbol{x}_{i}, y_{i}\right)~|~y_{i}=c\right\}$.\vspace*{0.15em}
		\State $\mathcal{Y}^{c} = \left\{{f_{\boldsymbol{\theta}}(\boldsymbol{x}_{i})}~|~\left(\boldsymbol{x}_{i}, y_{i}\right) \in \mathcal{D}^{c}\right\}$.\vspace*{0.15em}
		\State $\mathcal{G}^{c} = \text{\textsc{GradientUpperBounds}}\left(\mathcal{D}^{c}, \mathcal{Y}^{c}\right)$.~~~~~\verb+\\ using Eq. 7+\vspace*{0.15em}
		\If {$t \geq l$}\vspace*{0.15em}
		\State $\mathrm{LID}^{c} = \text{\textsc{GetLID}}\left(\mathcal{Y}^{c}, \mathrm{neighbors}=N\right)$.\vspace*{0.15em}
		\State $\mathrm{LID}^{c} = \text{\textsc{MovingAverage}}\left(\mathrm{LID}^{c}, \mathrm{window}=w\right)$.\vspace*{0.15em}
		\State $m=\left(1-k\right)\left|\mathcal{D}^{c}\right|/\left(T-l\right)$.\vspace*{0.15em}
		\State $idx = \text{\textsc{TopKargmax}}\left(\mathrm{LID}^{c}, \mathrm{K}=m\right)$.\vspace*{0.15em}
		\State Remove data with indices $idx$ from $\mathcal{D}^{c}$ and $\mathcal{D}$.\vspace*{0.15em}
		\State $\mathcal{G}^{c} = \text{\textsc{UpdateCoeffs}}\left(\mathcal{G}^{c}, \lambda, \mathrm{LID}^{c}\right)$.~~~~~\verb+\\ using Eq. 6+\vspace*{0.15em}
		\EndIf
		\State $S_{c}^{t}=\text{\textsc{GreedySolver}}\left(\mathcal{D}^{c}, \mathcal{G}^{c}, \mathrm{coreset~size}=k\right)$.\vspace*{0.15em}
		\State $S^{t}=S^{t} \cup S_{c}^{t}$.
		\EndFor
		\State Update neural network parameters~${\boldsymbol{\theta}}$ using stochastic gradient descent on~$S^{t}$.
		\EndFor
	\end{algorithmic}
\end{algorithm}

%-------------------------------------------------------------------------
\section{Implementation Details}\label{sec:ap:imp_det}
In this section, we provide the details of our experiments including hyperparameters, model architectures, and backdoor poisoning settings.
Note that all of the experiments were run using a single NVIDIA Tesla V100-SXM2-16GB GPU.

\begin{figure}[h!]
	\centering
	\begin{subfigure}{.26\textwidth}
		\centering
		\includegraphics[width=0.9\textwidth]{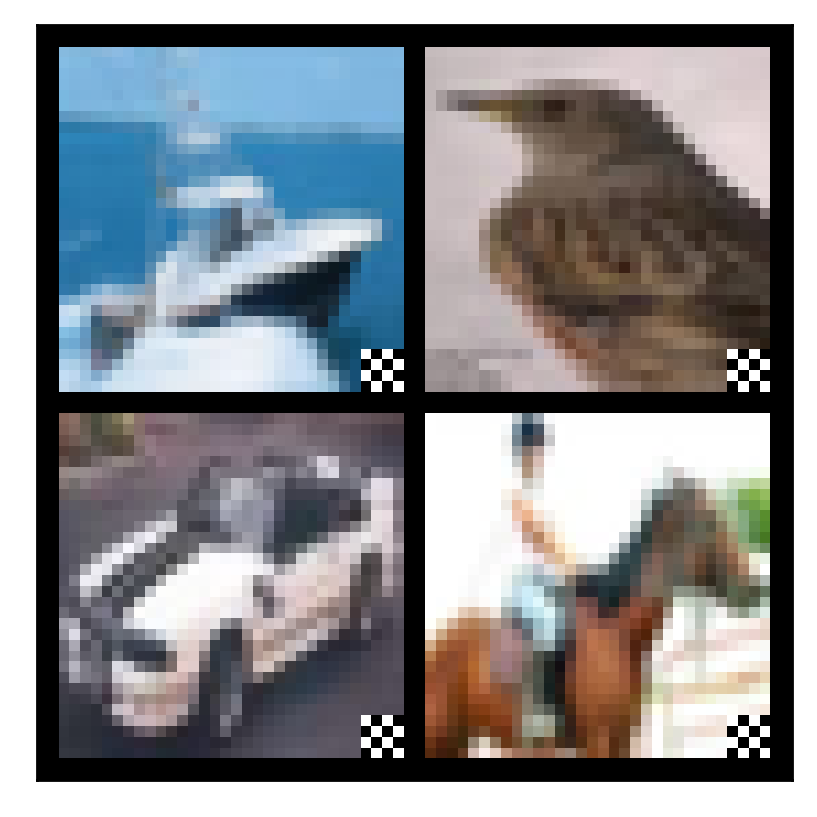}
		\caption*{BadNets}
	\end{subfigure}\hspace*{-0.45em}
	\begin{subfigure}{.26\textwidth}
		\centering
		\includegraphics[width=0.9\textwidth]{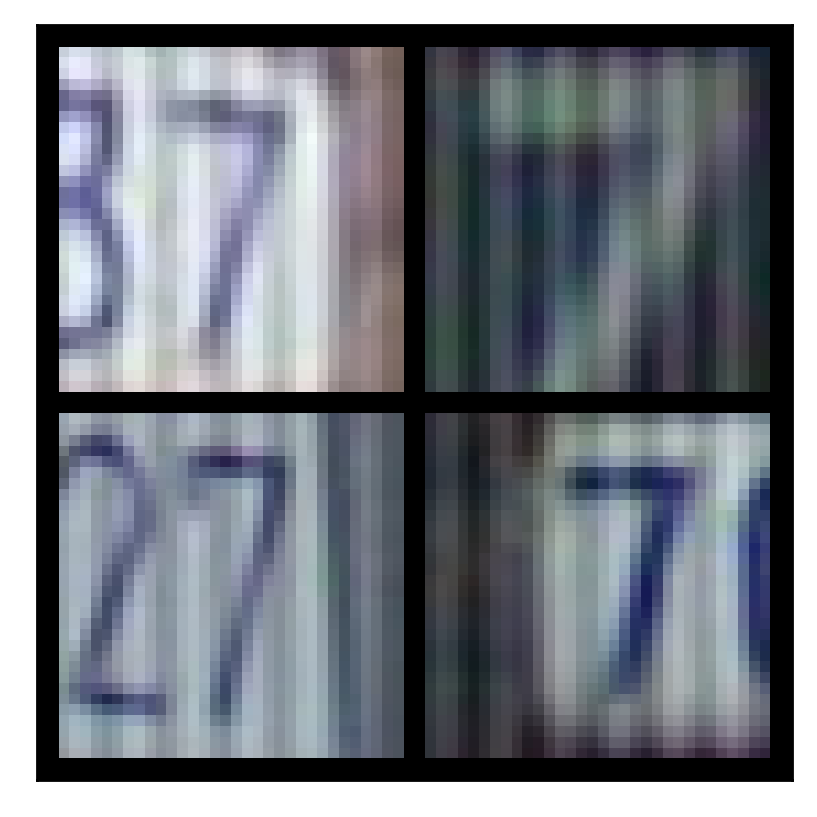}
		\caption*{Sinusoidal Strips}
	\end{subfigure}\hspace*{-0.45em}
	\begin{subfigure}{.26\textwidth}
		\centering
		\includegraphics[width=0.9\textwidth]{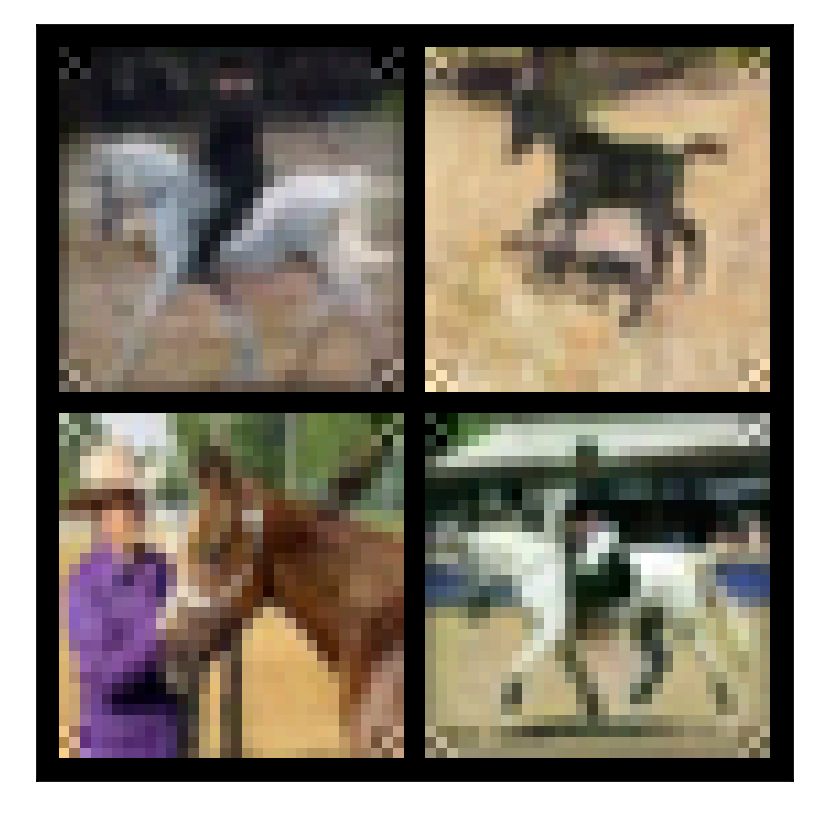}
		\caption*{Label-consistent}
	\end{subfigure}\hspace*{-0.45em}
	\begin{subfigure}{.26\textwidth}
		\centering
		\includegraphics[width=0.9\textwidth]{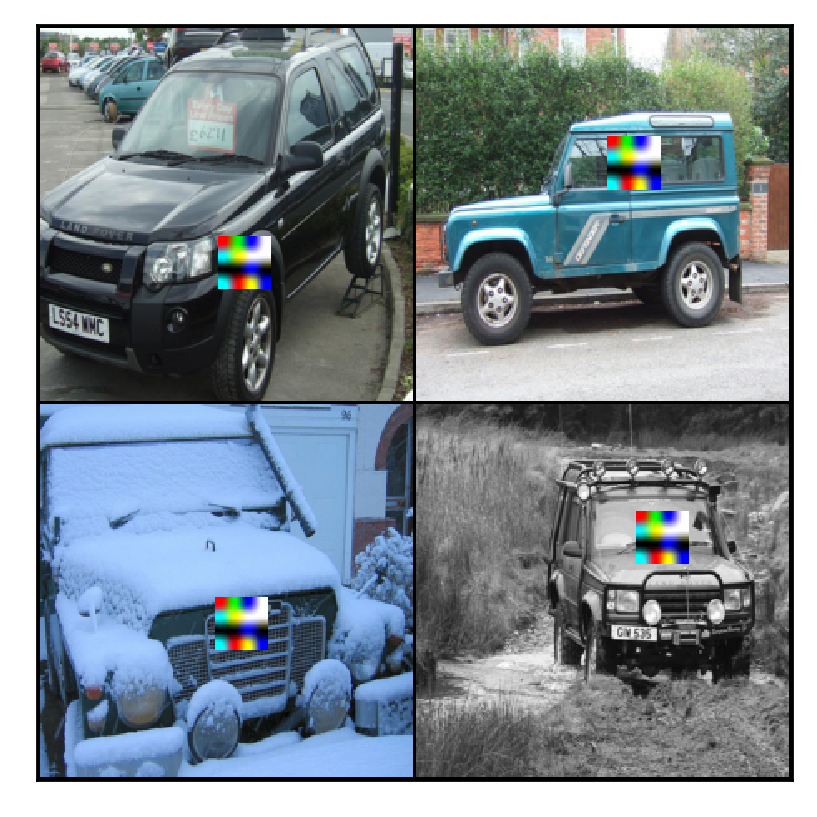}
		\caption*{HTBA}
	\end{subfigure}
	\caption{Samples of backdoor data poisonings used in experiments.}
	\label{fig:data_samples}
	\vskip -0.1in
\end{figure}

\paragraph{Datasets and Backdoor Poisoning Settings.}
We used CIFAR-10~\cite{krizhevsky2009learning}, SVHN~\cite{netzer2011reading}, and 12 randomly selected classes of ImageNet~\cite{russakovsky2015imagenet} for our experiments.
We padded CIFAR-10 and SVHN datasets with 4 zero pixels added to both sides of the image, and then randomly cropped each instance such that the final images become of size 32$\times$32.
For ImageNet-12, the zero-padding was done via 28 pixels, and the final image size after random cropping was set to 224$\times$224.
In each case, we randomly selected a \textit{target class}, and poisoned a fraction of the training data in that class with their backdoor counterpart.
The ratio of the poisoned examples in the target class is denoted as the \textit{injection rate}.
After injecting the poisoned data, we randomly chose a portion of the training data as our clean held-out validation set.
This data was used for model selection.
Finally, we used BadNets~\cite{gu2017badnets} with checkerboard pattern, label-consistent attacks~\cite{turner2019label-consistent}, sinusoidal strips~\cite{barni2019sig}, and HTBA triggers~\cite{saha2020htba} as our backdoor data poisoning rules.\footnote{For label-consistent attacks~\cite{turner2019label-consistent}, we used the poisoned data provided by~\href{https://github.com/MadryLab/label-consistent-backdoor-code}{the official repository}. In particular, we used the adversarial data generated with $\ell_{2}$-bounded perturbations, where the bound is set to 300.}
Samples of each poisoned data can be found in \Cref{fig:data_samples}.
\Cref{tab:data} summarizes the settings of each dataset used in our experiments.

\paragraph{Model Architecture \& Training Hyperparameters.}
We used stochastic gradient descent (SGD) optimizers in our experiments.
The momentum and weight decay for all datasets were set to 0.9 and 5e-4, respectively.
For CIFAR-10 and SVHN datasets, we train ResNet-32~\cite{he2016deep} models for 120 epochs.
The initial learning rate was set to 0.1, which was later divided by 10 at epochs 80 and 100.
For ImageNet-12, ResNet-18~\cite{he2016deep} neural networks were trained for 200 epochs.
In this case, the initial learning rate was also set to 0.1, and it was later divided by 10 at epochs 72 and 144.
For training with coresets, the ratio of the data selected from each class is denoted as the \textit{coreset size}.
Moreover, for \textsc{Collider}, the epoch at which the LID is enabled is denoted as the \textit{LID start epoch}.
For the LID term in \textsc{Collider} (\Cref{eq:collider}), a Lagrange multiplier is also required.
This hyperparameter, denoted by $\lambda$, was set to 0.01 after tuning for CIFAR-10 dataset, and kept fixed through the other experiments.
\Cref{tab:hyper} shows all the hyperparameters used for training the neural networks in our experimental study.

\begin{table*}[htp]\setlength{\tabcolsep}{3.pt}
	\caption{Details of the datasets used in our experiments.}
	\vskip -0.1in
	\label{tab:data}
	\begin{center}
		\begin{tiny}
			\begin{tabular}{ccccccc}
				\toprule
				Dataset          & Image Size      & Zero-padding & Backdoor        & Target Class     & Injection Rate & Val. Data Ratio \\
				\midrule
				CIFAR-10         & 32$\times$32    & 4            & BadNets         & Airplane         & 10\%           & 4\% \\
				CIFAR-10         & 32$\times$32    & 4            & Label Consist.  & Horse            & 10\%           & 4\% \\        
				SVHN             & 32$\times$32    & 4            & Sin. Strips     & Digit 7          & 10\%           & 4\% \\ 
				ImageNet-12      & 224$\times$224  & 28           & HTBA Triggers   & Jeep, land-rover & 40\%           & 20\% \\ 
				\bottomrule
			\end{tabular}
		\end{tiny}
	\end{center}
	\vskip -0.1in
\end{table*}

\begin{table*}[htp]
	\caption{Training hyperparameters used in our experiments.}
	\label{tab:hyper}
	
	\begin{center}
		\begin{small}
			\begin{tabular}{lcc}
				\toprule
				\multirow{2}{*}{Hyperparameter}    & \multicolumn{2}{c}{Dataset}\\
				\cmidrule{2-3}
				& CIFAR-10 \& SVHN                 & ImageNet-12\\
				\midrule
				Optimizer                          & SGD                              & SGD\\
				Scheduler                          & Multi-step                       & Multi-step\\
				Initial lr.                        & 0.1                              & 0.1\\
				lr. decay (epochs)                 & 10 (80, 100)                     & 10 (72, 144)\\
				Batch Size                         & 128                              & 32\\
				Epochs                             & 120                              & 200\\
				\midrule
				Model arch.                        & ResNet-32                        & ResNet-18\\
				\midrule
				Coreset size                       & \multicolumn{2}{c}{0.4 for Sin. Strips, 0.3 otherwise}\\
				LID Start Epoch                    & \multicolumn{2}{c}{30 for BadNets and Sin. Strips, 50 for Label-consist. and HTBA}\\
				$\lambda$                          & \multicolumn{2}{c}{0.01}\\
				\bottomrule
			\end{tabular}
		\end{small}
	\end{center}
	\vskip -0.1in
\end{table*}

%-------------------------------------------------------------------------
\section{Extended Experimental Results}\label{sec:ap:ext_sim_res}
In this section, we present our extended simulation results.

\paragraph{Performance Measures.}
We use three measures to compare the performance of each training algorithm on the backdoor poisoned data.
First, we compute the accuracy of each model on the clean test set, and denote it with ACC.
Next, we evaluate the robustness of each method by the attack success rate (ASR).
To compute this quantity, we first remove all the data that originally belong to the target class.
Then, we install the trigger used in each case to poison the rest of the test set.
We then compute the accuracy of the poisoned data, which should lead the model to output the target class.
Finally, we evaluate the purity of our coresets.
To this end, we define a quantity which we call the \textit{filtered poison data}.
In a nutshell, this measure shows the portion of the poisoned data that is left out of the coreset.
This value is always between zero and one.
Zero means that all the poisoned data is in our selected coreset.
In contrast, one shows that the selected coreset is free of any poisoned data.
To ensure reproduciblity, each experiment is repeated with 5 different random seeds.\footnote{Our implementation can be found in \href{https://github.com/hmdolatabadi/COLLIDER}{this repository}.} 
Unless specified otherwise, in each case we report the mean alongside an errorbar which is the standard deviation across these 5 seeds.
This errorbar is shown by a shaded area or bars in plots. 

\paragraph{ImageNet-12 Results.}
\Cref{tab:large_result} shows our experimental results for ImageNet-12 dataset.
Like all the previous small size image datasets, \textsc{Collider} gives the best robustness against backdoor data poisonings in this case.

\begin{table*}[htp!]
	\caption{Clean test accuracy (ACC) and attack success rate (ASR) in \% for HTBA~\cite{saha2020htba} backdoor triggers on ImageNet12~\cite{russakovsky2015imagenet} dataset.
		The results show the mean and standard deviation for 5 different seeds.
		In this case, 40\% of the data in the target class contains backdoor poisoned data.}
	\vskip -0.1in
	\label{tab:large_result}
	\begin{center}
		\begin{small}
			\begin{tabular}{ccc}
				\toprule
				\multirow{2}{*}{Training}                                               & \multicolumn{2}{c}{HTBA~\cite{saha2020htba}}\\
				\cmidrule(lr){2-3} 
				& ACC          & ASR\\
				\midrule
				Vanilla                                                                 & $91.43 \pm 0.52$	& $49.36 \pm 14.42$\\
				Coresets                                                                & $87.29 \pm 0.50$  & $37.63 \pm 4.84$ \\ 
				\textsc{Collider}                                                       & $85.15 \pm 0.32$	& $19.11 \pm 3.05$\\
				\bottomrule
			\end{tabular}
		\end{small}
	\end{center}
	\vskip -0.1in
\end{table*}

\paragraph{WANet~\cite{nguyen2021wanet} Results.}
As suggested by the reviewers, we run a similar experiment to \Cref{tab:large_result} over CIFAR-10 dataset that was poisoned with WANet~\cite{nguyen2021wanet}.
As seen in \Cref{tab:wanet}, our approach leads to a significant reduction of the attack success rate in this state-of-the-art attack.
Note that for this case, we just ran our algorithm without any rigorous hyperparameter tuning.

\begin{table*}[tb!]\setlength{\tabcolsep}{4.5pt}
	\caption{Clean test accuracy (ACC) and attack success rate (ASR) in \% for WANet~\cite{nguyen2021wanet} data poisonings on CIFAR-10.
		The results show the mean and standard deviation for 5 different seeds.
		The poisoned data injection rate is 40\%.
		In this case, the coreset size is 0.4.}
	\label{tab:wanet}
	\begin{center}
		\begin{footnotesize}
			\begin{tabular}{ccc}
				\toprule
				\multirow{2}{*}{Training}                     & \multicolumn{2}{c}{WANet~[36]}\\
				\cmidrule(lr){2-3}
				                                              & ACC                & ASR\\
				\midrule
				Vanilla                                       & $91.63 \pm 0.28$   & $92.24 \pm 1.74$ \\ 
				Coresets                                      & $86.04 \pm 0.89$   & $5.73  \pm 2.78$ \\ 
				\textsc{Collider}                             & $84.27 \pm 0.55$   & $4.29  \pm 2.54$  \\
				\bottomrule
			\end{tabular}
		\end{footnotesize}
	\end{center}
\end{table*}

\paragraph{Semi-supervised Learning on Non-coreset Data.}
Instead of excluding non-coreset data from the training process, one can add them as a set of unlabeled data and exploit semi-supervised learning to train the model.
To this end, we use MixMatch~\cite{berthelot2019mixmatch} as our semi-supervised learning approach.\footnote{We use the PyTorch implementation of MixMatch available \href{https://github.com/YU1ut/MixMatch-pytorch}{here}.} 
Moreover, we update the neural network weights using an exponential moving average, and change the optimizer to Adam~\cite{kingma2015adam} to comply with the MixMatch implementation that we use.
Apart from these tweaks, all the \textsc{Collider} hyperparameters were kept similar to their original settings.

We perform semi-supervised learning~(SSL) on the coreset selection as well as \textsc{Collider}.
To differentiate the effect of semi-supervised learning from coreset selection, we also randomly select a fraction of the data and remove their labels during each epoch and run MixMatch~\cite{berthelot2019mixmatch} on them.

\Cref{tab:ssl_results} shows our results using semi-supervised learning.
For reference, our results without using MixMatch is also shown at the first three rows of \Cref{tab:ssl_results}.
As seen, using MixMatch we can improve the clean accuracy gap with the vanilla training on BadNets~\cite{gu2017badnets} and Label-consistent~\cite{turner2019label-consistent} attacks.
This use, however, has an adverse effect on Sinusoidal Strips~\cite{barni2019sig}.

\begin{table*}[htp]\setlength{\tabcolsep}{4.5pt}
	\caption{Clean test accuracy (ACC) and attack success rate (ASR) in \% for backdoor data poisonings on CIFAR-10~(BadNets and label-consistent) and SVHN (sinusoidal strips) datasets.
		The results show the mean and standard deviation for 5 different seeds.
		The poisoned data injection rate is 10\%.
		For BadNets and label-consistent attacks, the coreset size is 0.3. It is 0.4 (0.2) for sinusoidal strips (+SSL).}
	\vskip -0.1in
	\label{tab:ssl_results}
	\begin{center}
		\begin{tiny}
			\begin{tabular}{ccccccc}
				\toprule
			    \multirow{2}{*}{Training}    & \multicolumn{2}{c}{BadNets~\cite{gu2017badnets}}  & \multicolumn{2}{c}{Label-consistent~\cite{turner2019label-consistent}}  & \multicolumn{2}{c}{Sinusoidal Strips~\cite{barni2019sig}} \\
				\cmidrule(lr){2-3} \cmidrule(lr){4-5} \cmidrule(lr){6-7}
			                                                                          	& ACC               & ASR               & ACC               & ASR             & ACC               & ASR\\
				\midrule
				Vanilla                                                                 & $92.19 \pm 0.20$	& $99.98 \pm 0.02$	& $92.46 \pm 0.16$	& $100$           & $95.79 \pm 0.20$  & $77.35 \pm 3.68$\\
				Coresets                                                                & $84.86 \pm 0.47$  & $74.93 \pm 34.6$  & $83.87 \pm 0.36$  & $7.78 \pm 9.64$ & $92.30 \pm 0.19$  & $24.30 \pm 8.15$\\ 
				\textsc{Collider}                                                       & $80.66 \pm 0.95$	& $4.80 \pm 1.49$   & $82.11 \pm 0.62$	& $5.19 \pm 1.08$ & $89.74 \pm 0.31$  & $6.20 \pm 3.69$\\
				\midrule
				Random Subset + SSL                                                     & $91.32 \pm 0.35$	& $99.57 \pm 0.49$	& $91.48 \pm 0.16$	& $100$           & $95.71 \pm 0.10$  & $65.22 \pm 1.89$\\
				Coresets + SSL                                                          & $88.36 \pm 0.26$  & $4.45 \pm 0.34$   & $87.82 \pm 0.18$  & $6.56 \pm 2.87$ & $91.12 \pm 0.74$  & $42.62 \pm 3.88$\\ 
				\textsc{Collider} + SSL                                                 & $84.67 \pm 0.62$	& $4.10 \pm 0.58$   & $84.33 \pm 0.37$	& $2.62 \pm 0.22$ & $90.53 \pm 0.45$  & $31.17 \pm 7.69$\\
				\bottomrule
			\end{tabular}
		\end{tiny}
	\end{center}
	\vskip -0.15in
\end{table*}

\begin{figure}[htp]
	\centering
	\begin{subfigure}{\textwidth}
		\centering
		\includegraphics[width=0.475\textwidth]{LID_Norm_Average.png}
		\hfill
		\includegraphics[width=0.475\textwidth]{LID_Sample_Average.png}
		\caption{BadNets}
	\end{subfigure}
	\vskip\baselineskip
	\begin{subfigure}{\textwidth}
		\centering
		\includegraphics[width=0.475\textwidth]{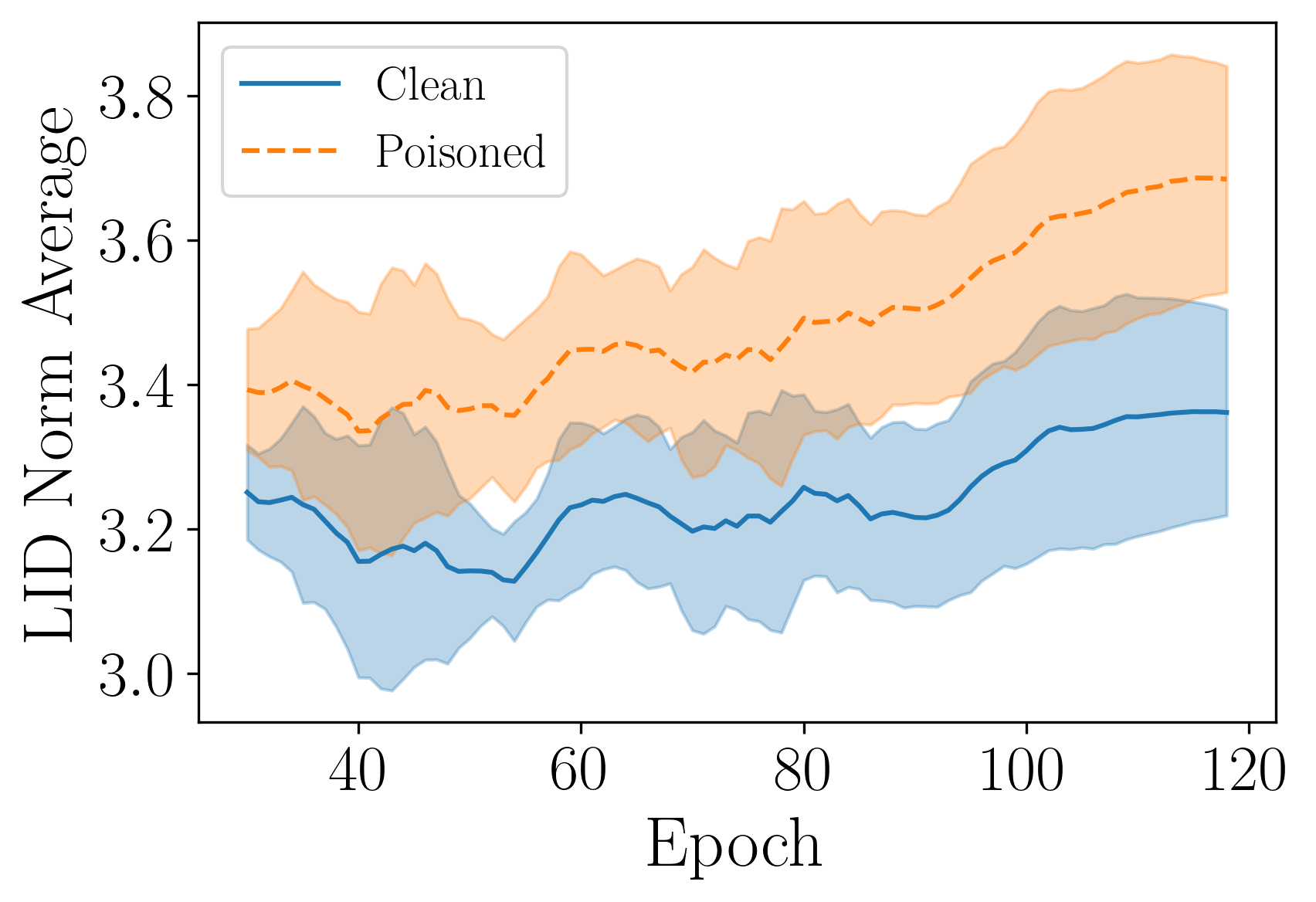}
		\hfill
		\includegraphics[width=0.475\textwidth]{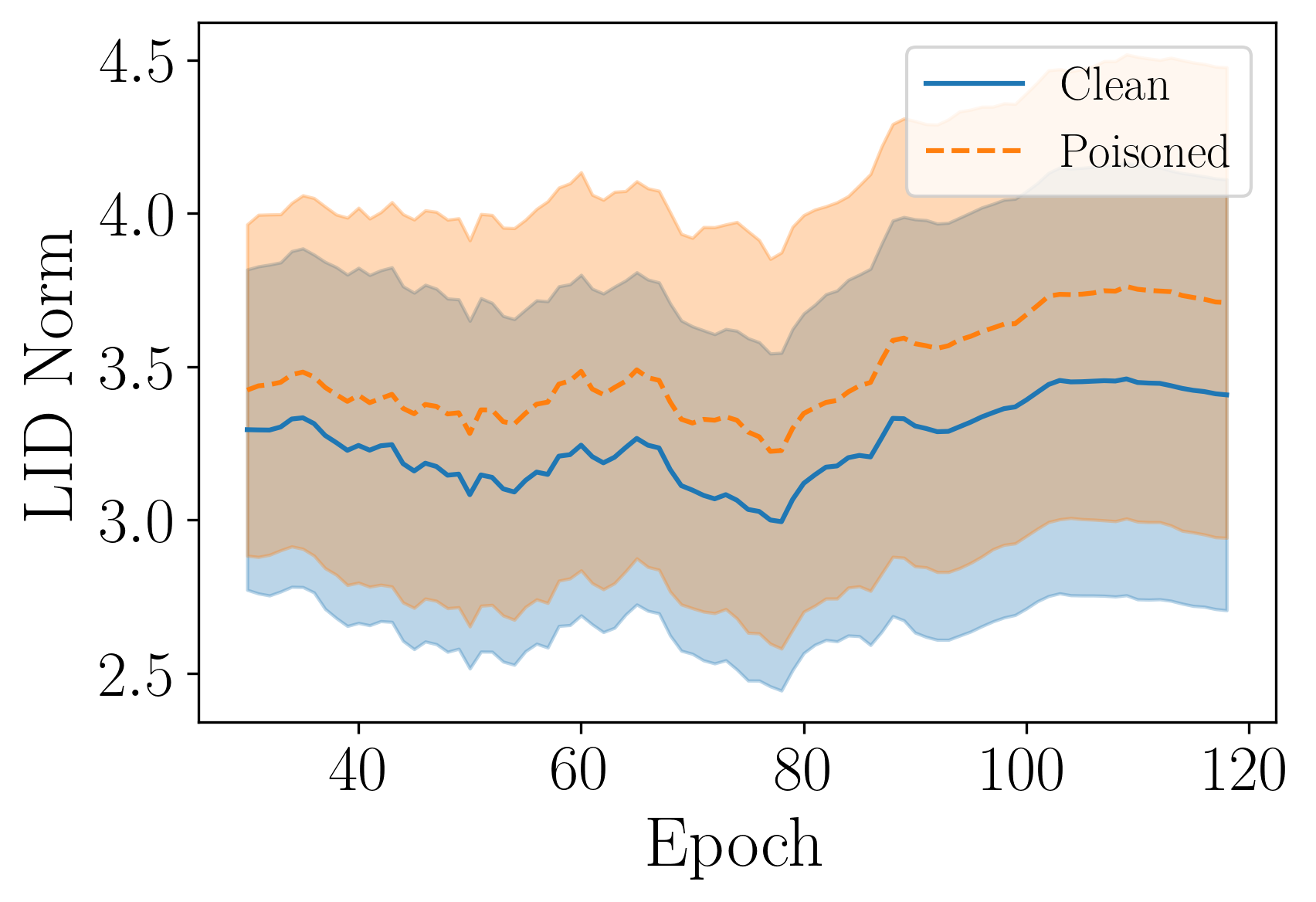}
		\caption{Label-consistent Attack}
	\end{subfigure}
	\vskip\baselineskip
	\begin{subfigure}{\textwidth}
		\centering
		\includegraphics[width=0.475\textwidth]{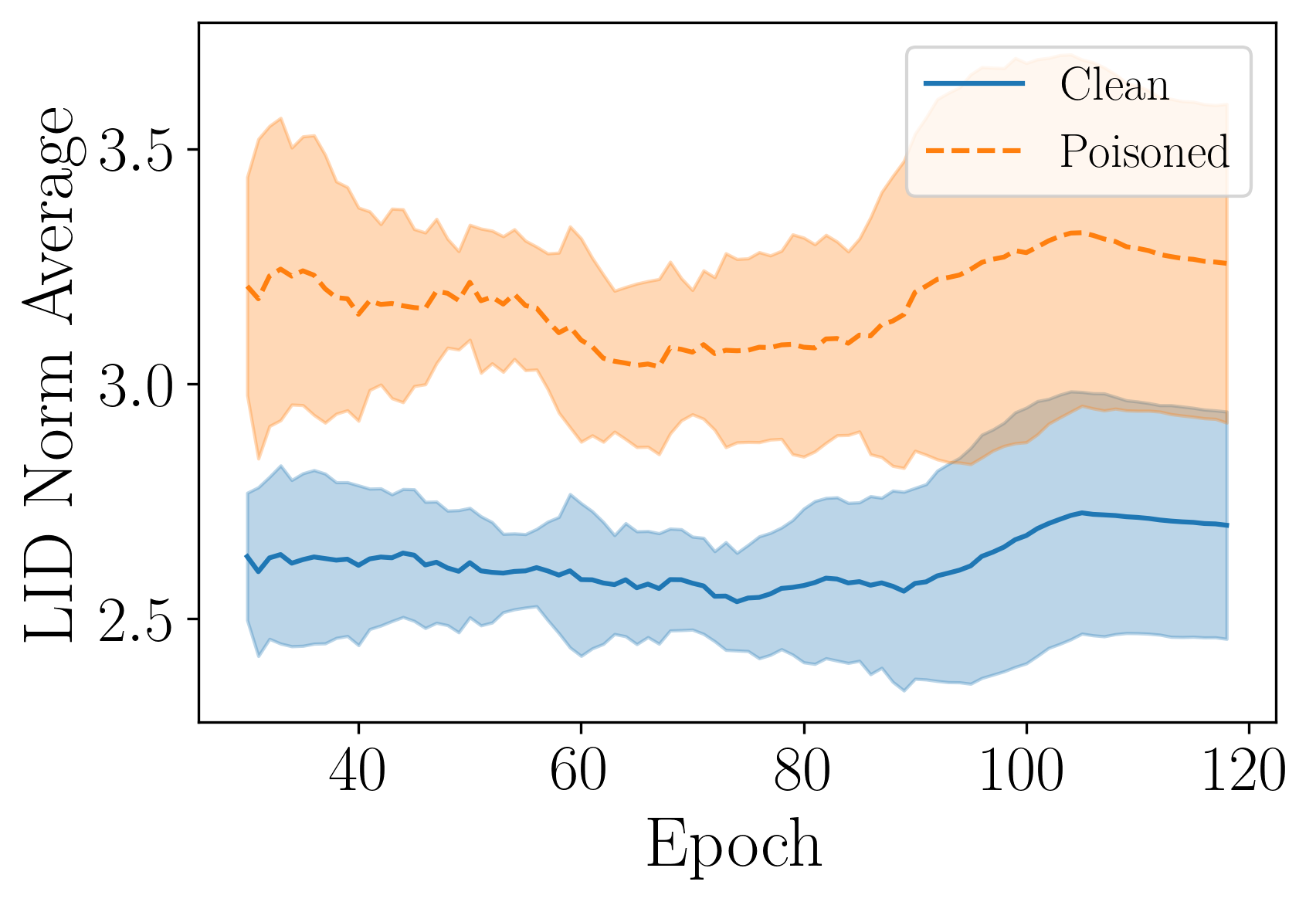}
		\hfill
		\includegraphics[width=0.475\textwidth]{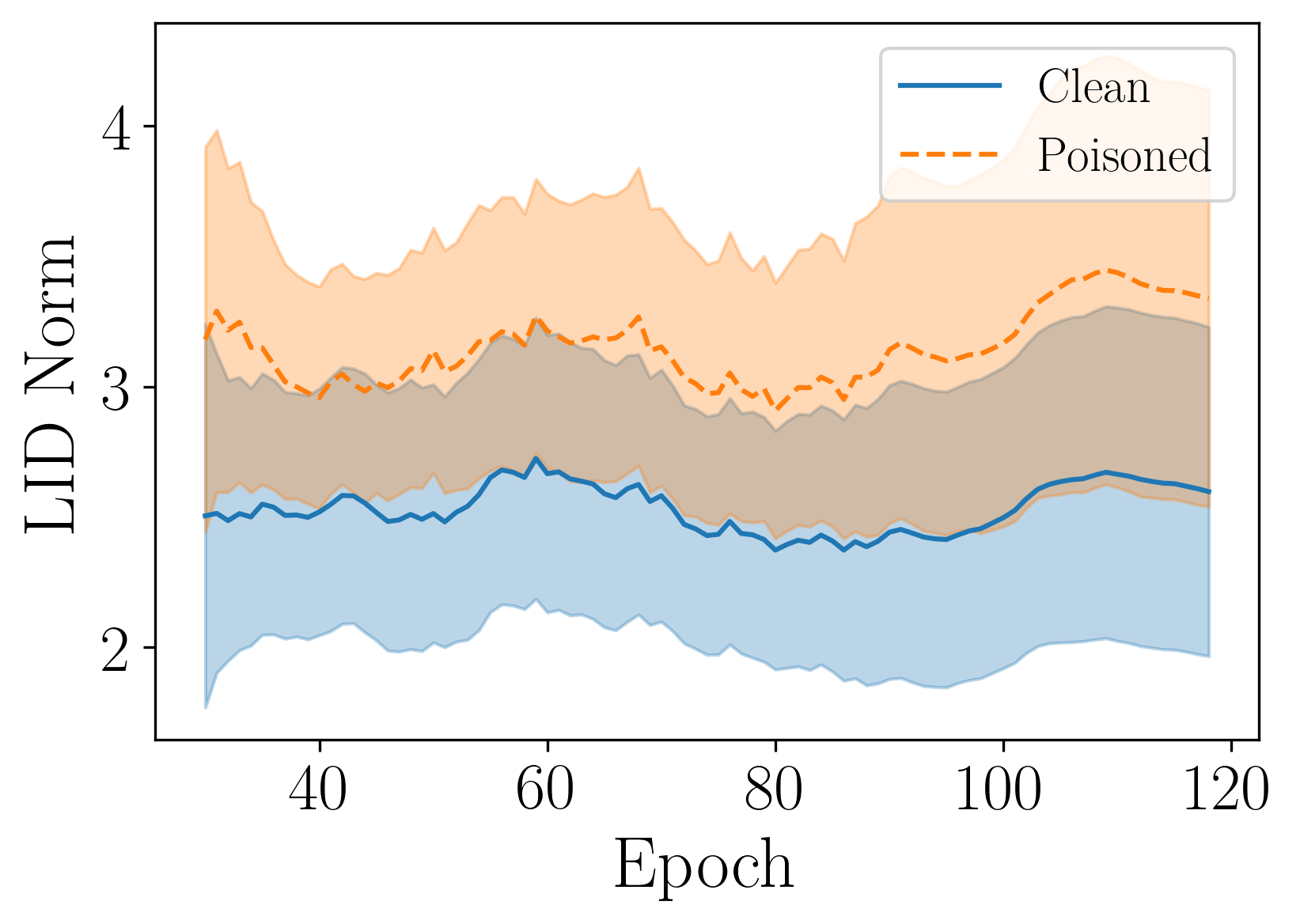}
		\caption{Sinusoidal Strips}
	\end{subfigure}
	\caption{The LID of clean and backdoor poisoned data samples.
		Left: average LID norm across 5 different seeds.
		Right: LID distribution for a single run.}
	\label{fig:ap:LIDs}
	\vskip -0.1in
\end{figure}

\paragraph{LID of Poisoned vs. Clean Data.}
\Cref{fig:ap:LIDs} shows the LID values for clean and poisoned data samples, which are poisoned by BadNets, label-consistent attacks, and sinusoidal strips, respectively.
As seen, there definitely is a difference between clean and poisoned data in terms of the local intrinsic dimensionality.
This difference becomes less severe for label-consistent attacks that use reduced-intensity triggers.
Still, the right tail of the LID distribution in all the cases consist of poisoned data only.
This observation justifies our choice of throwing them away permanently from the training set.

\begin{figure}[htp]
	\centering
	\begin{subfigure}{\textwidth}
		\centering
		\includegraphics[width=0.32\textwidth]{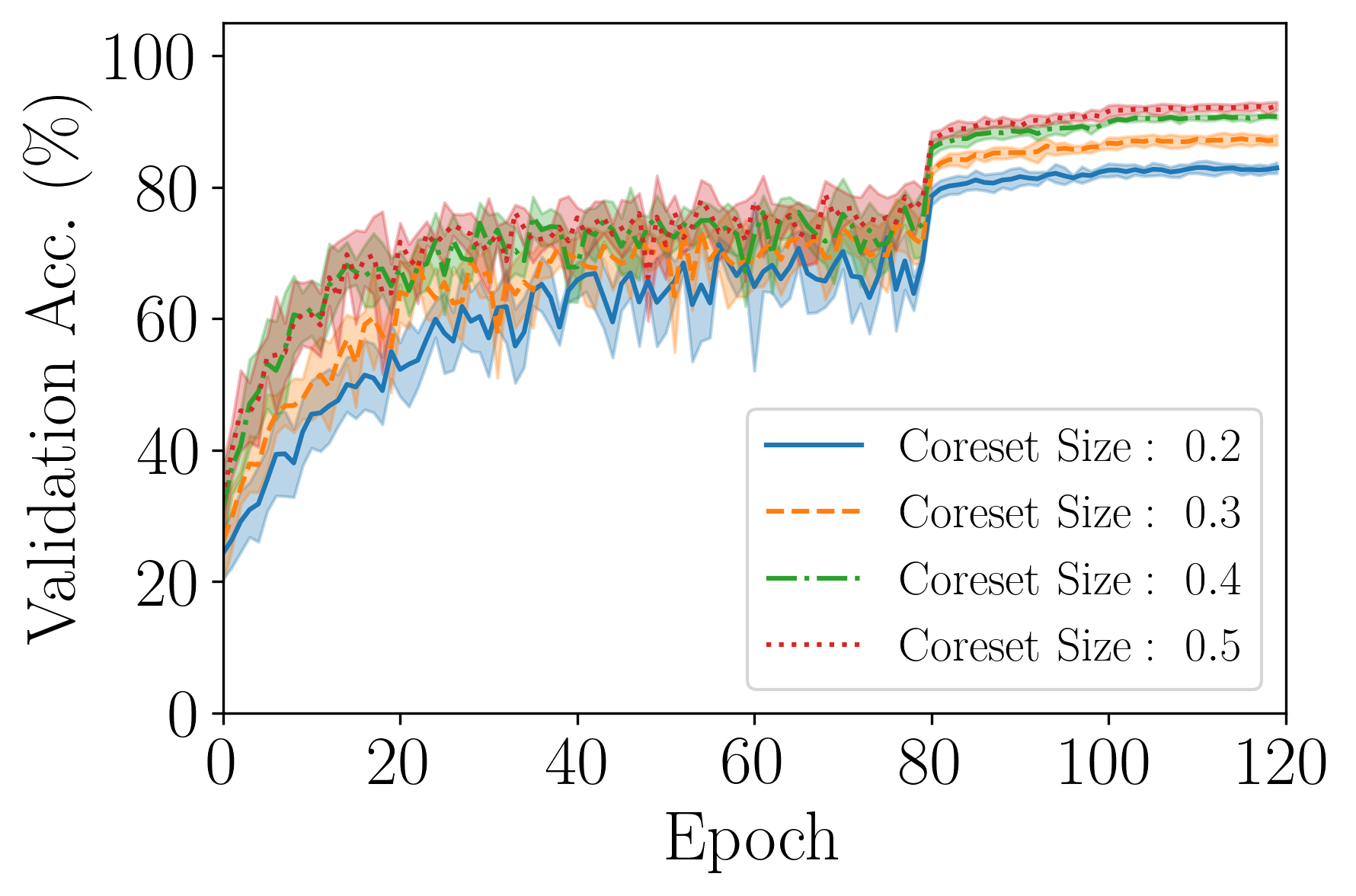}
		\hfill
		\includegraphics[width=0.32\textwidth]{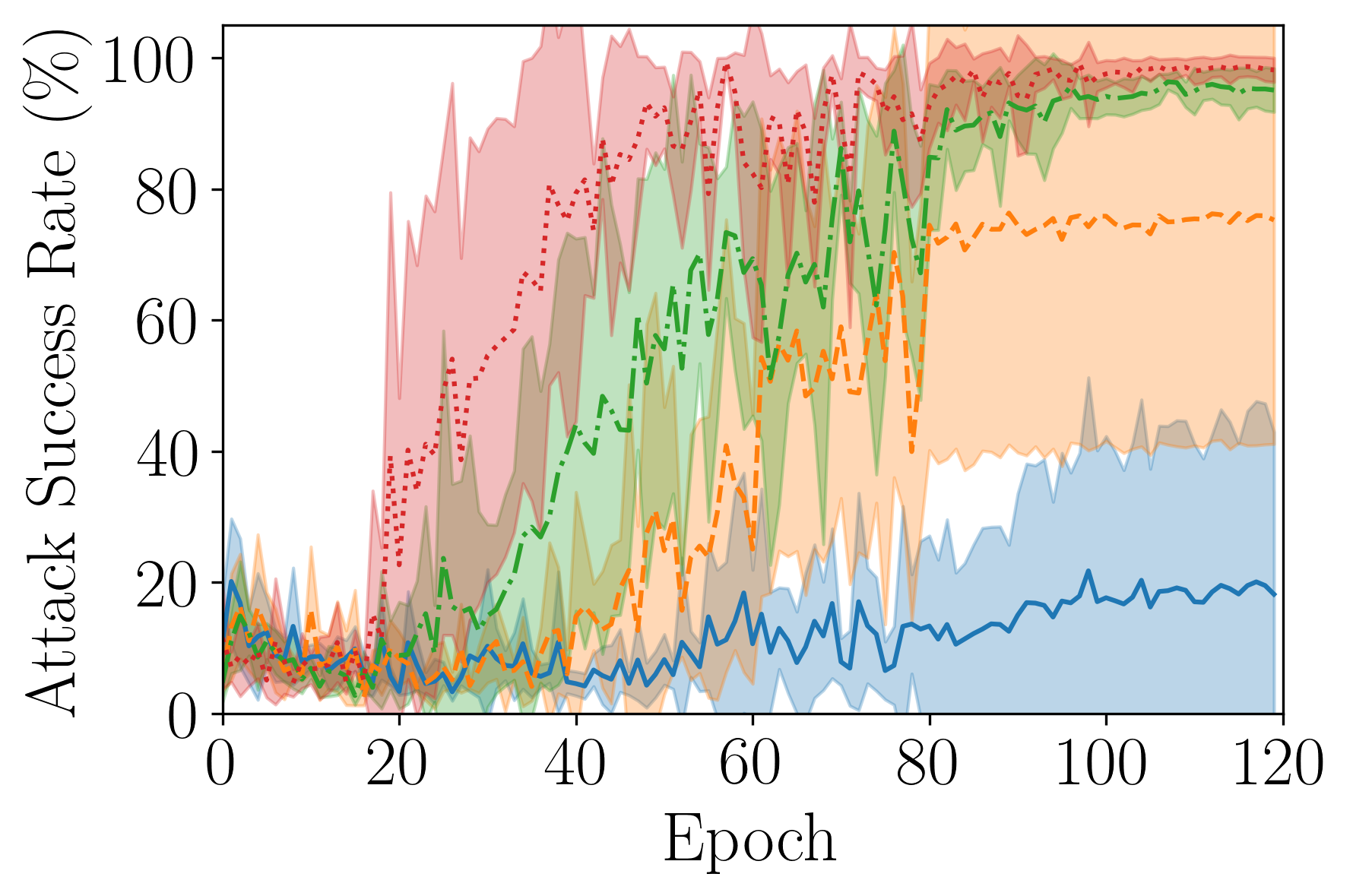}
		\hfill
		\includegraphics[width=0.32\textwidth]{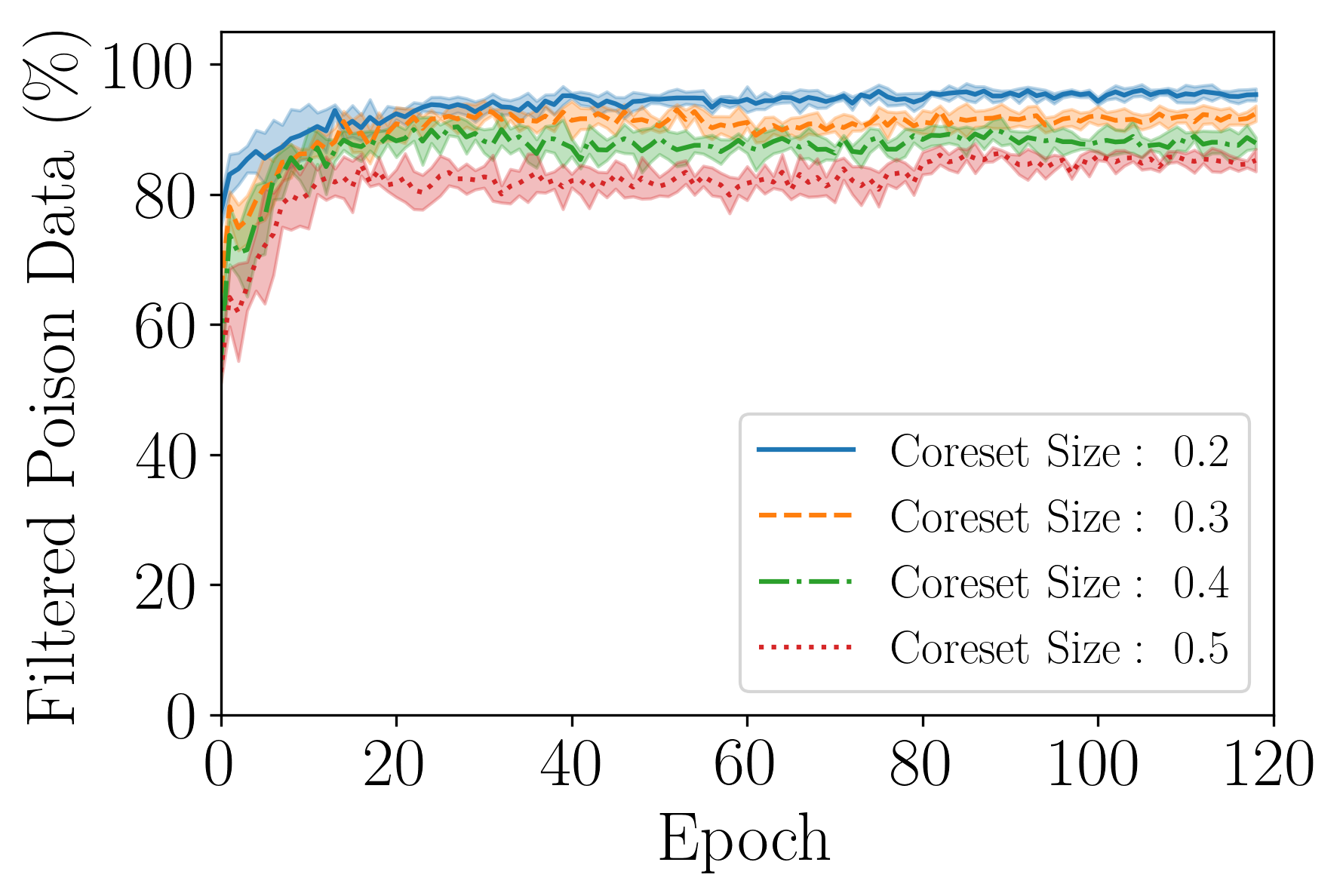}
		\caption{Base coreset selection~(\Cref{eq:basic_coreset})}
	\end{subfigure}
	\vskip\baselineskip
	\begin{subfigure}{\textwidth}
		\centering
		\includegraphics[width=0.32\textwidth]{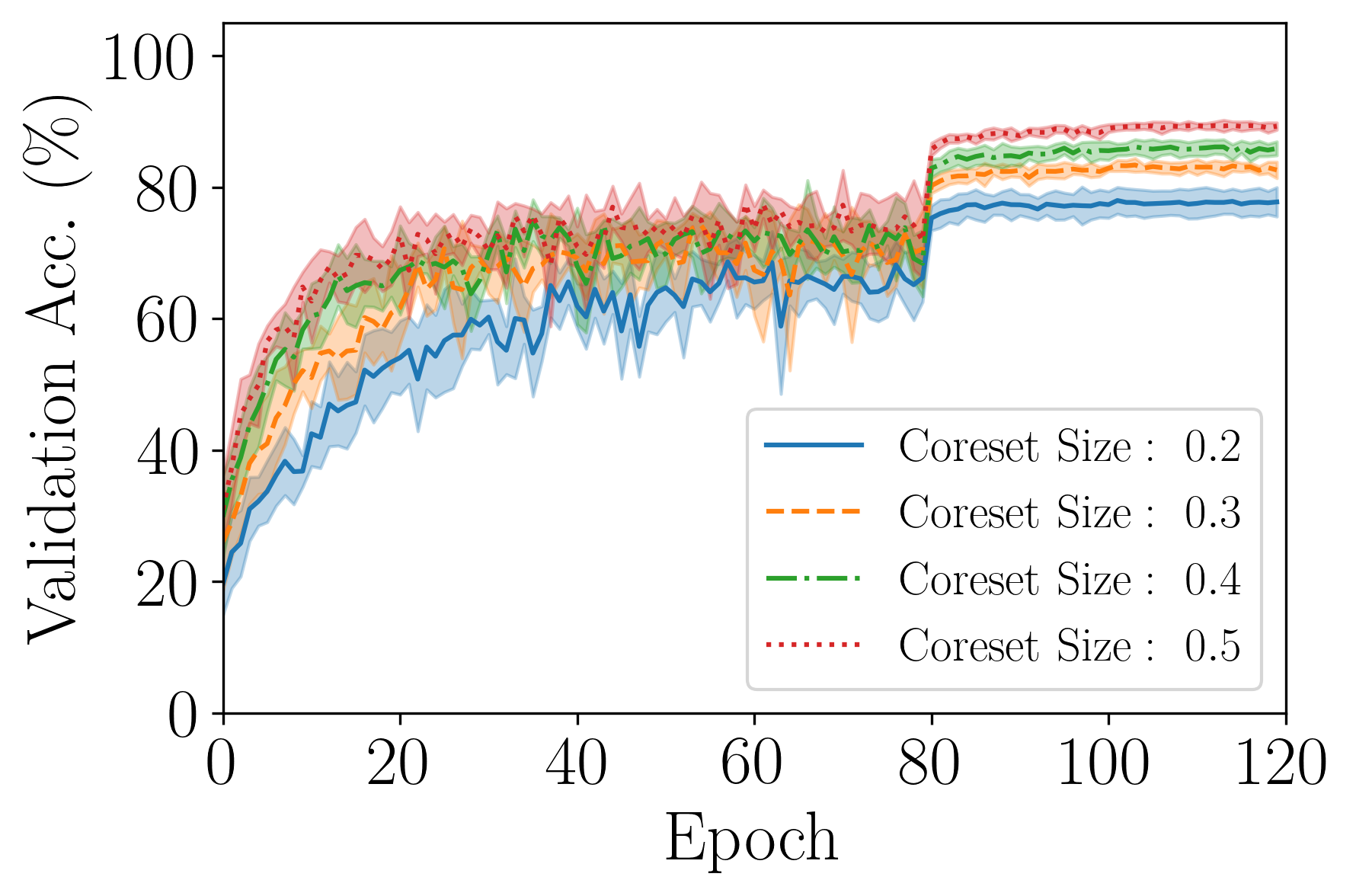}
		\hfill
		\includegraphics[width=0.32\textwidth]{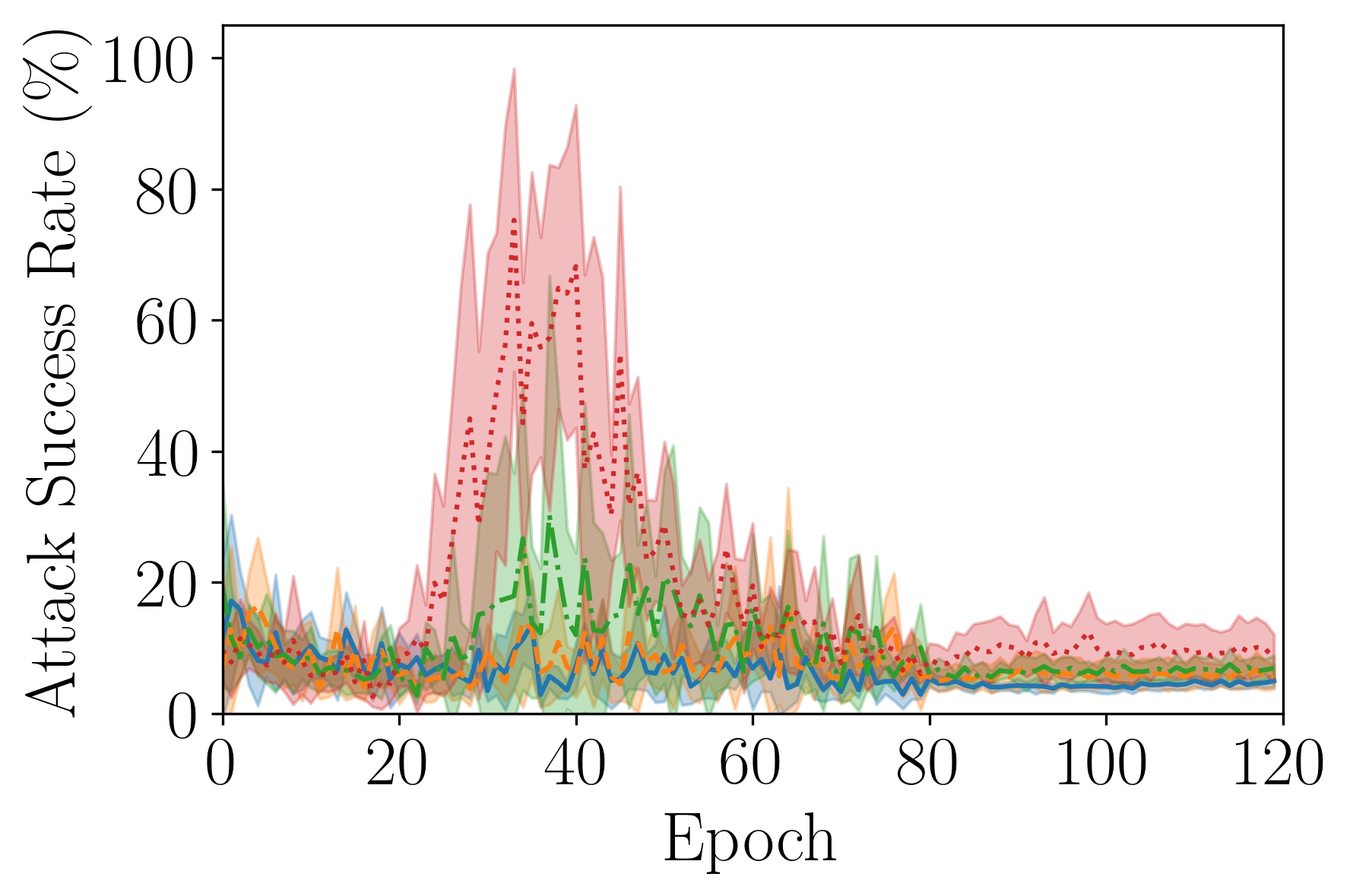}
		\hfill
		\includegraphics[width=0.32\textwidth]{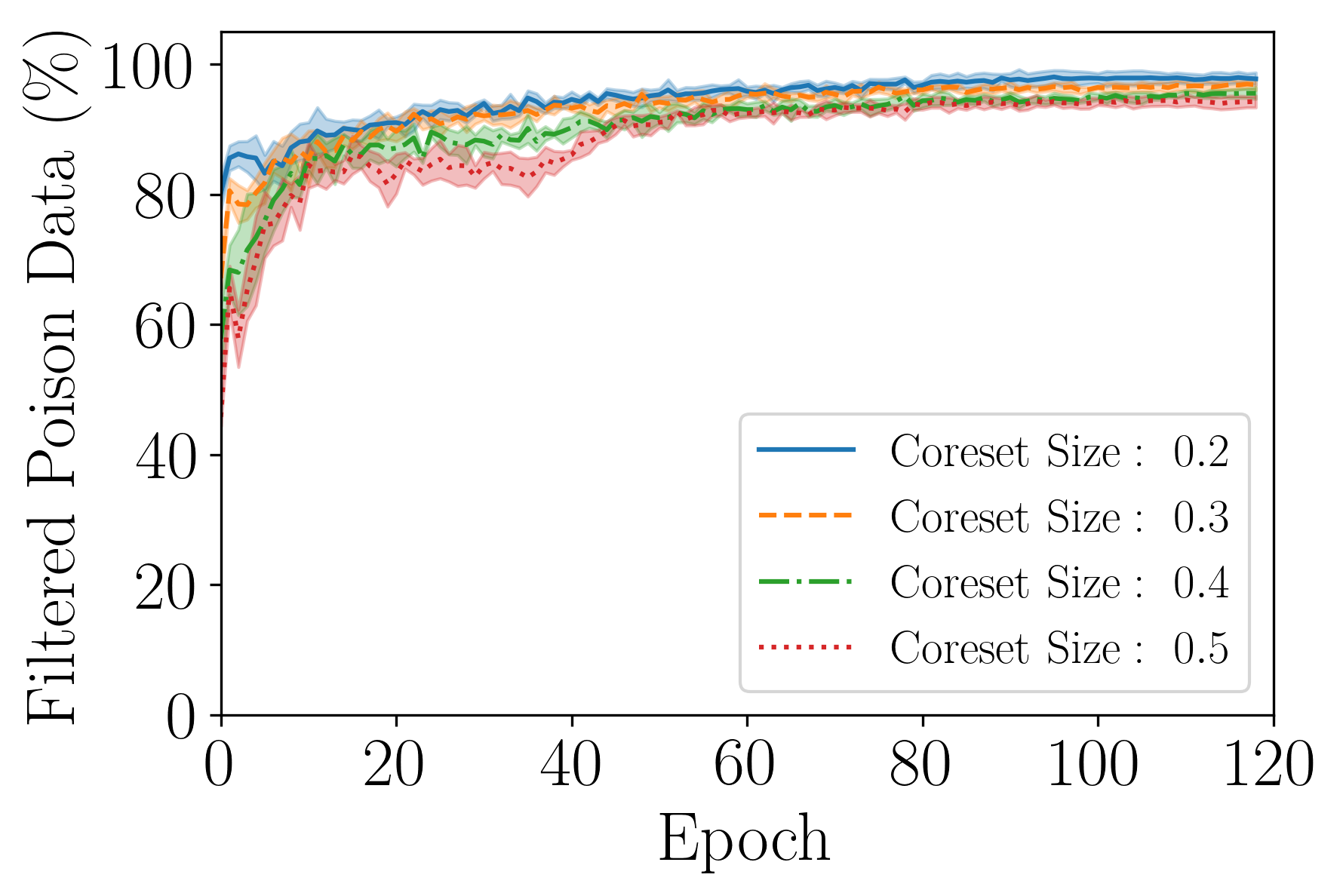}
		\caption{\textsc{Collider}}
	\end{subfigure}
	\caption{Evolution of the performance measures (validation accuracy, attack success rate, and filtered poison data) with training.
		The training dataset is CIFAR-10 which is poisoned by injecting 10\% backdoor data into its target class.
		(a) Basic gradient-based coreset selection
		(b) \textsc{Collider}: gradient-based coreset selection with LID regularization.} 
	\label{fig:ap:coreset_size}
	\vskip -0.1in
\end{figure}

\begin{figure}
	\centering
	\begin{subfigure}{\textwidth}
		\centering
		\includegraphics[width=0.4\textwidth]{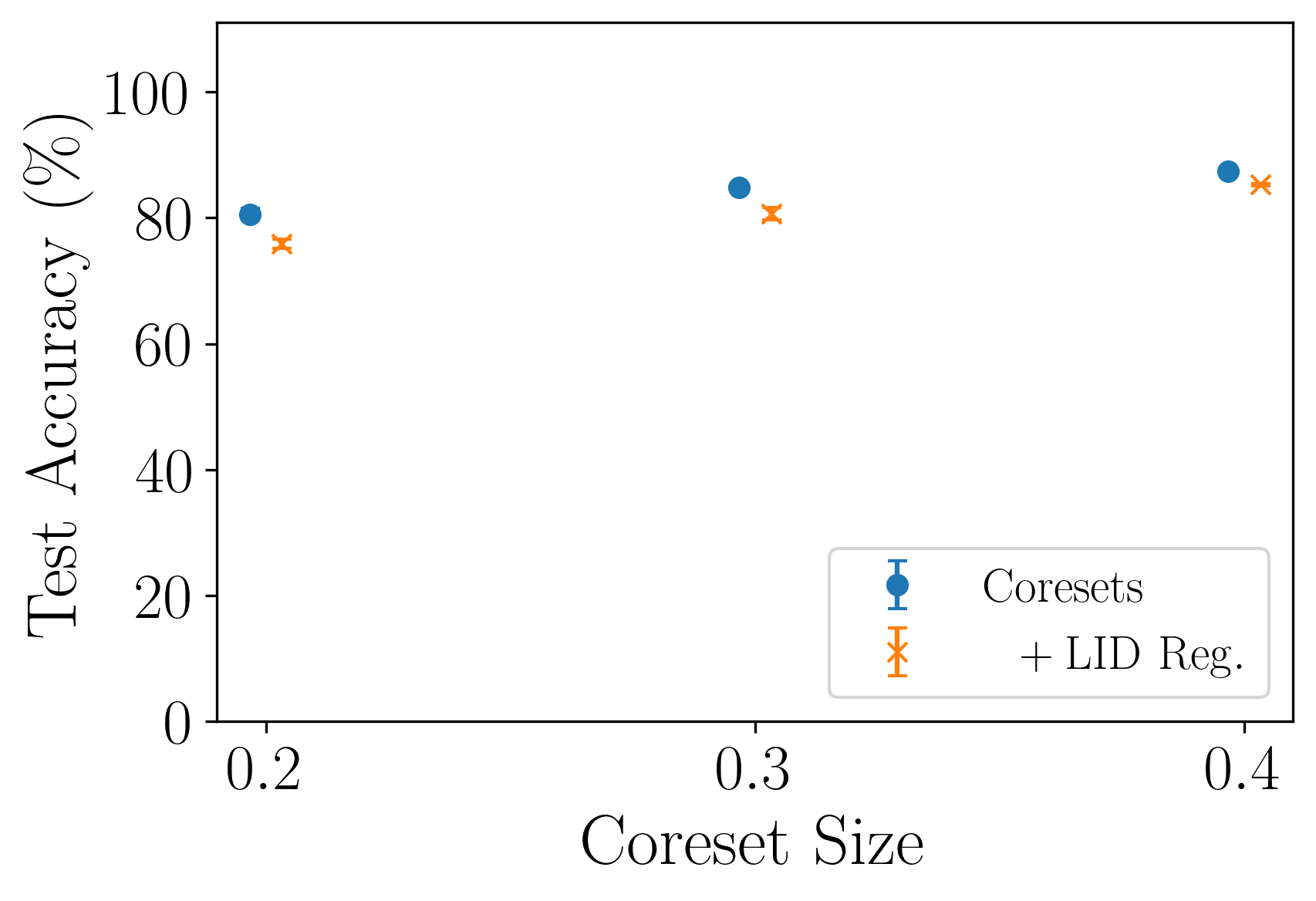}
		\hfill
		\includegraphics[width=0.4\textwidth]{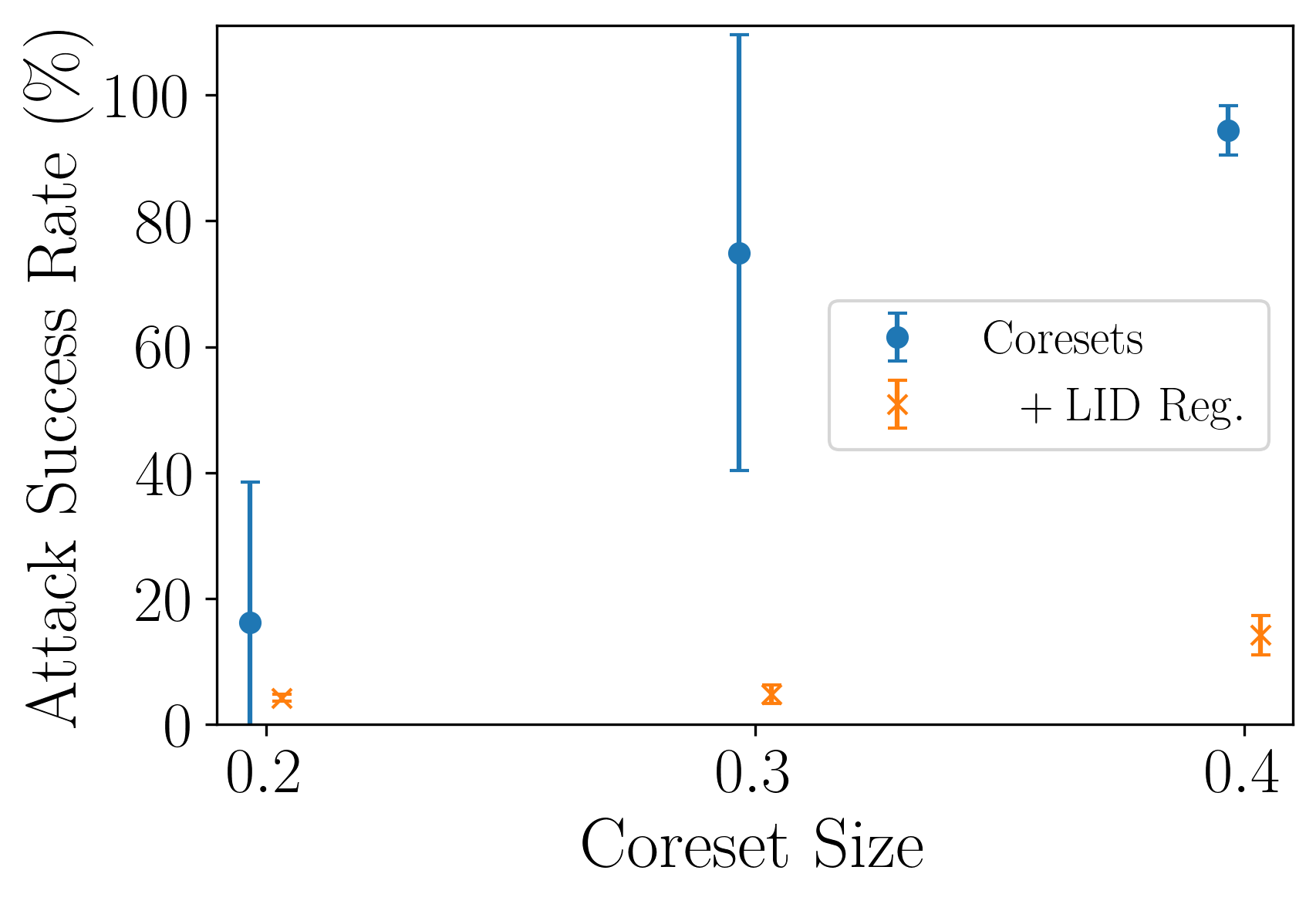}
		\caption{CIFAR-10, BadNets}
	\end{subfigure}
	\vskip\baselineskip
	\begin{subfigure}{\textwidth}
		\centering
		\includegraphics[width=0.4\textwidth]{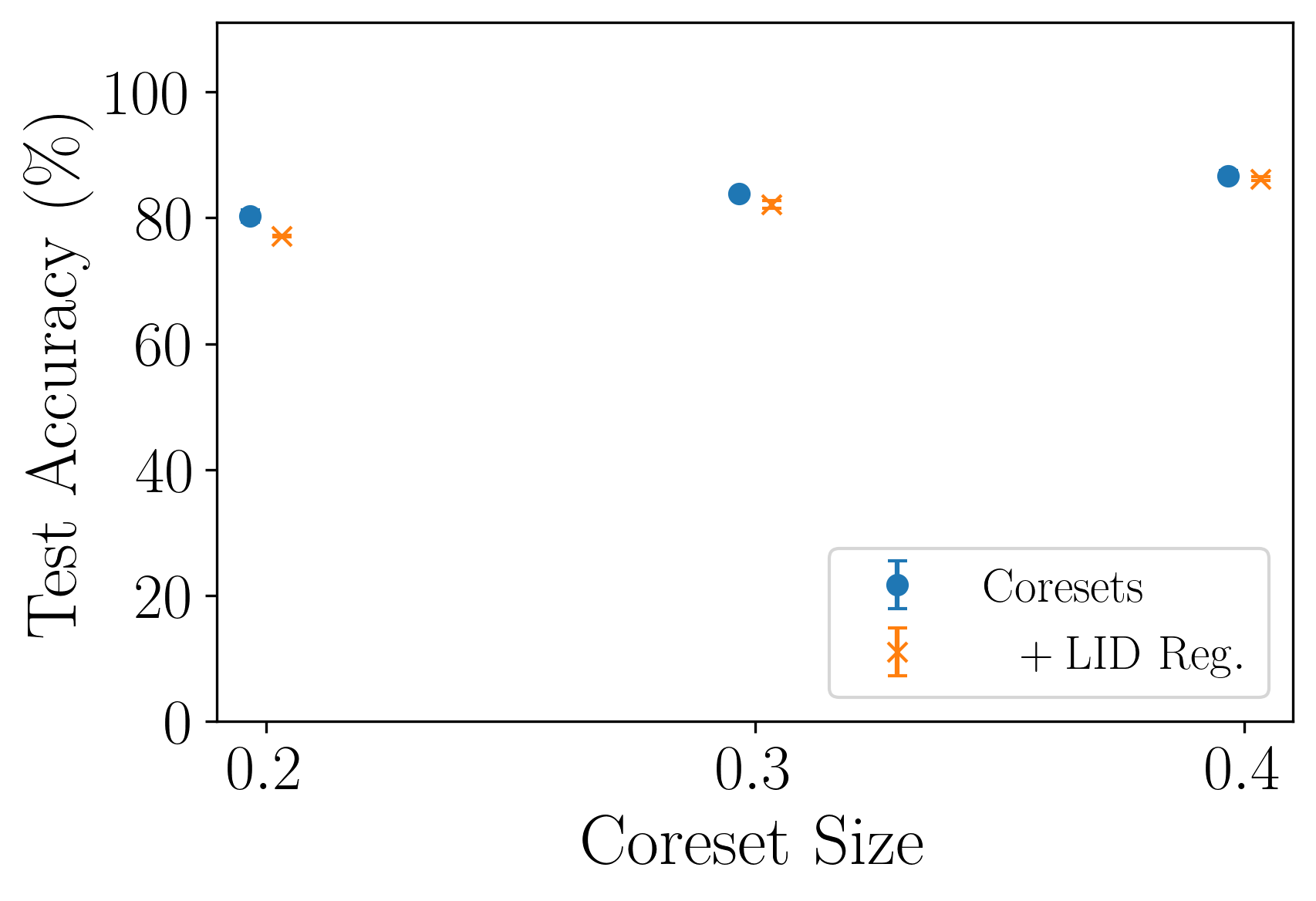}
		\hfill
		\includegraphics[width=0.4\textwidth]{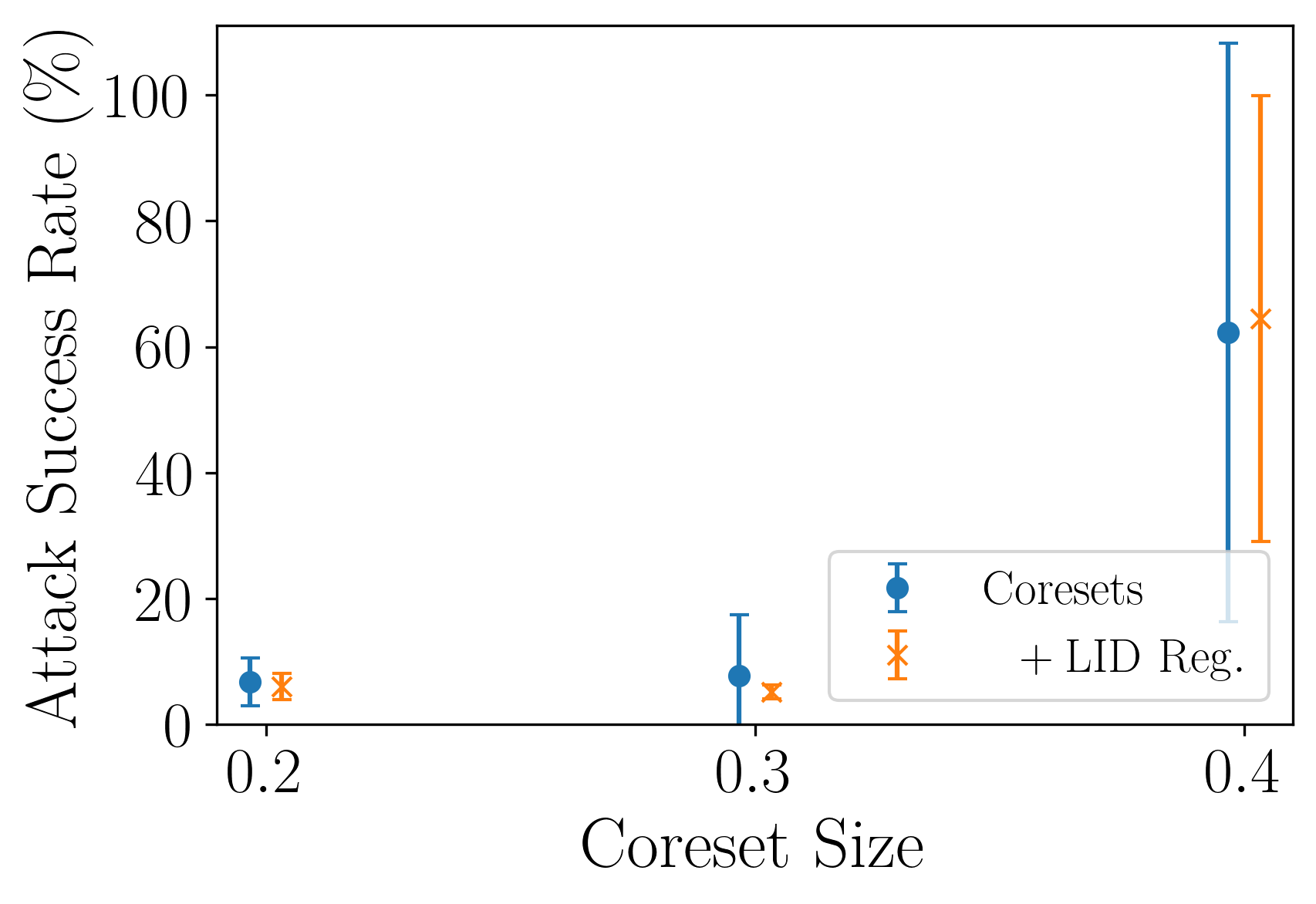}
		\caption{CIFAR-10, Label-consistent Attacks}
	\end{subfigure}
	\vskip\baselineskip
	\begin{subfigure}{\textwidth}
		\centering
		\includegraphics[width=0.4\textwidth]{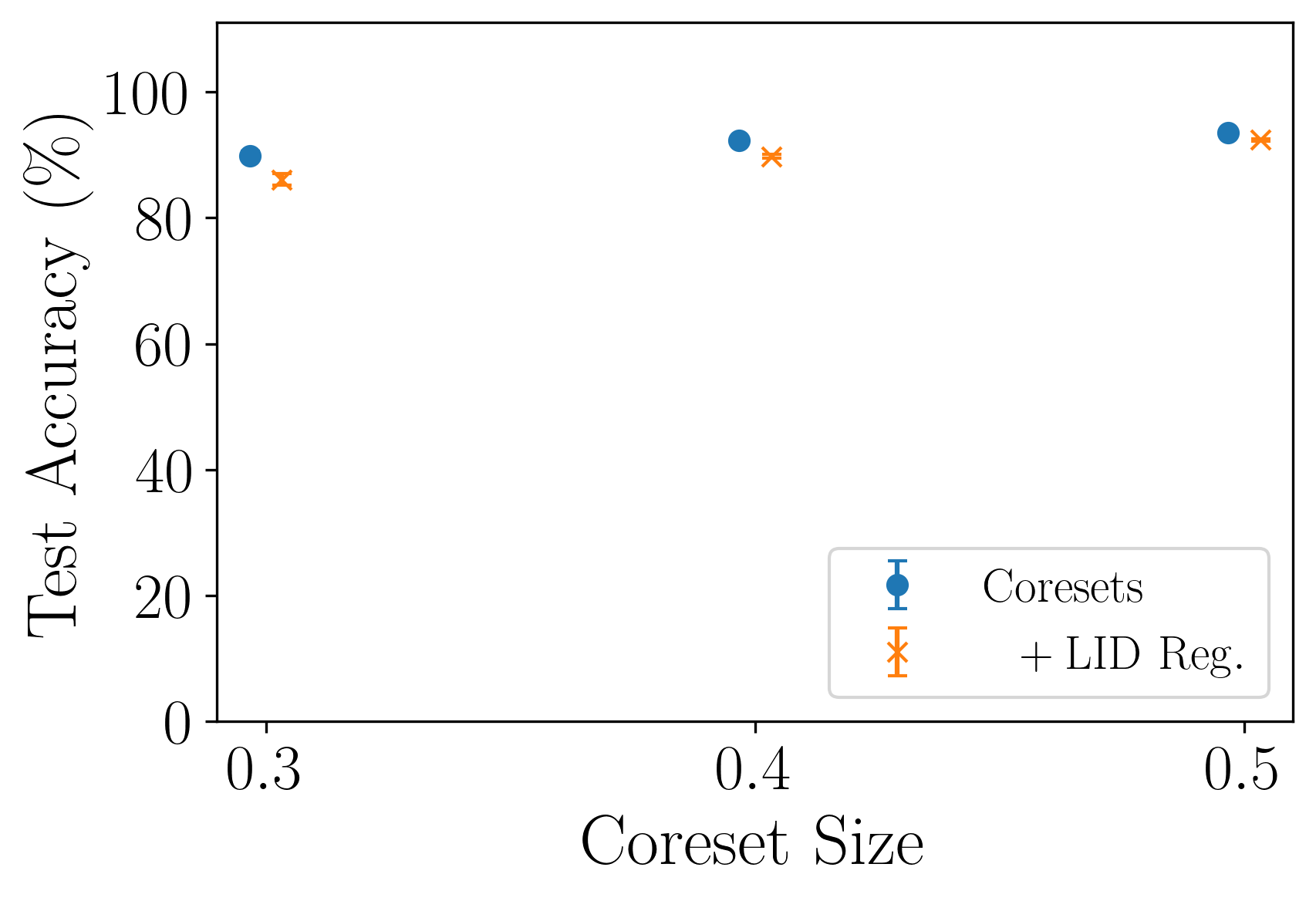}
		\hfill
		\includegraphics[width=0.4\textwidth]{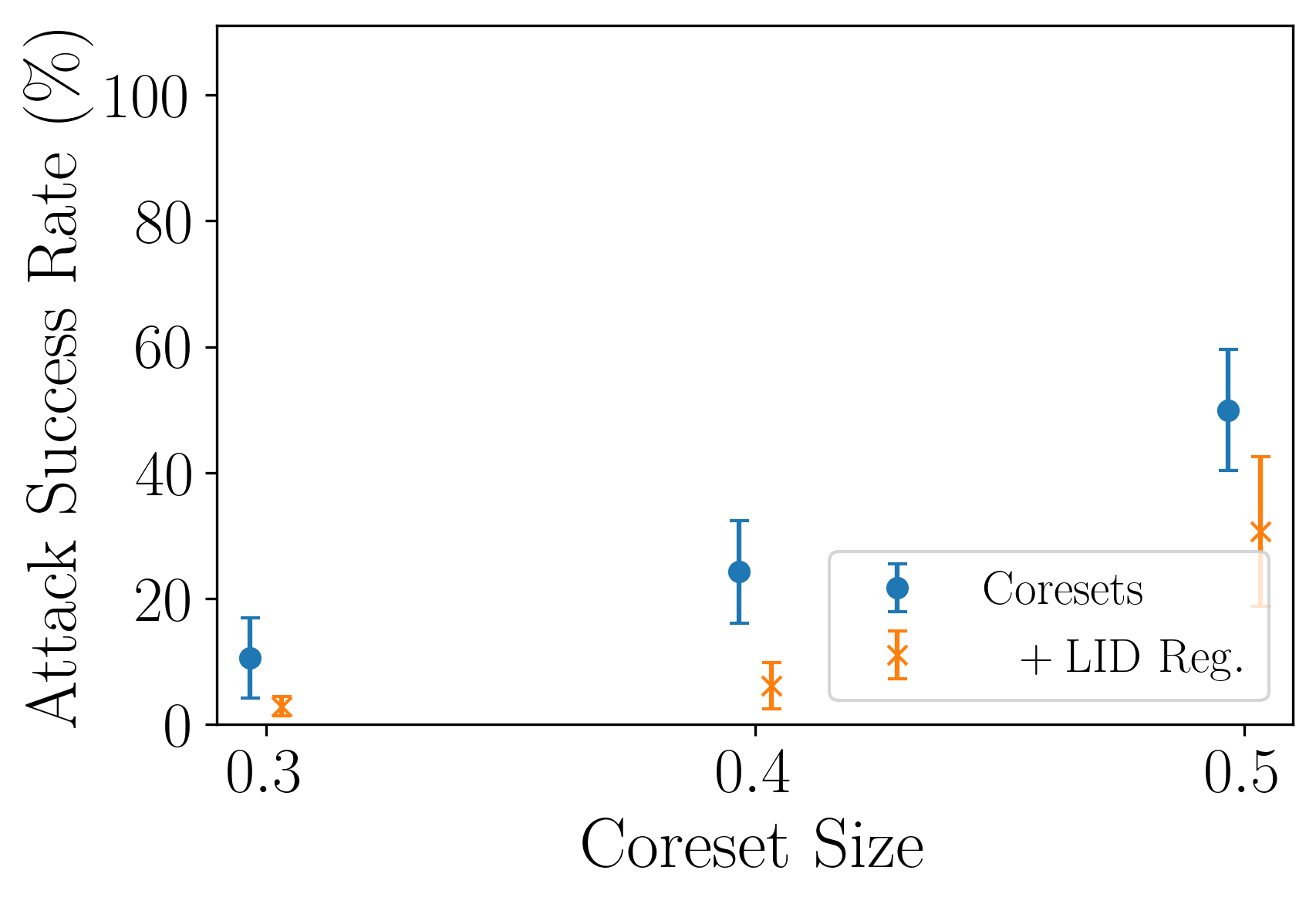}
		\caption{SVHN, Sinusoidal Strips}
	\end{subfigure}
	\vskip\baselineskip
	\begin{subfigure}{\textwidth}
		\centering
		\includegraphics[width=0.4\textwidth]{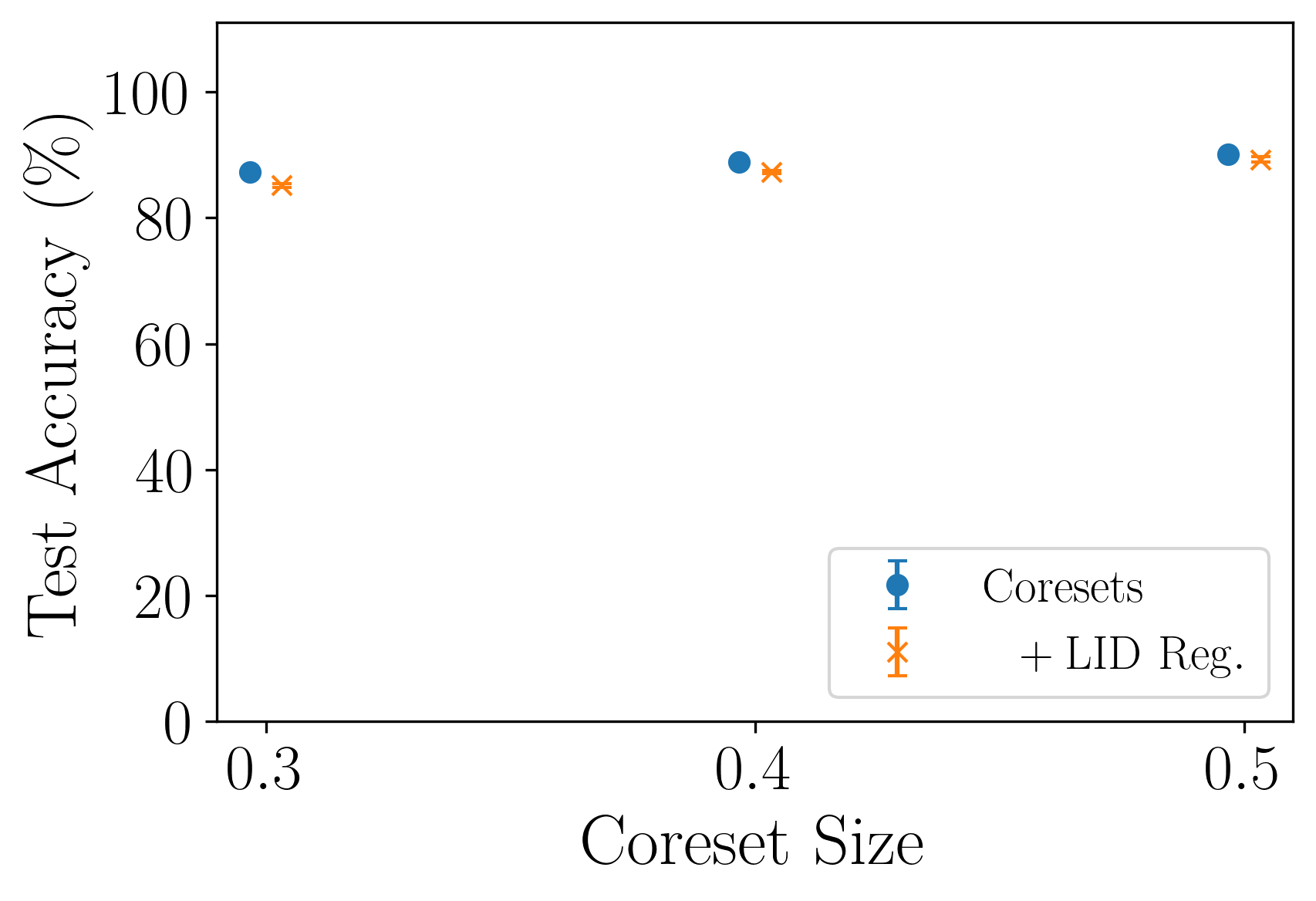}
		\hfill
		\includegraphics[width=0.4\textwidth]{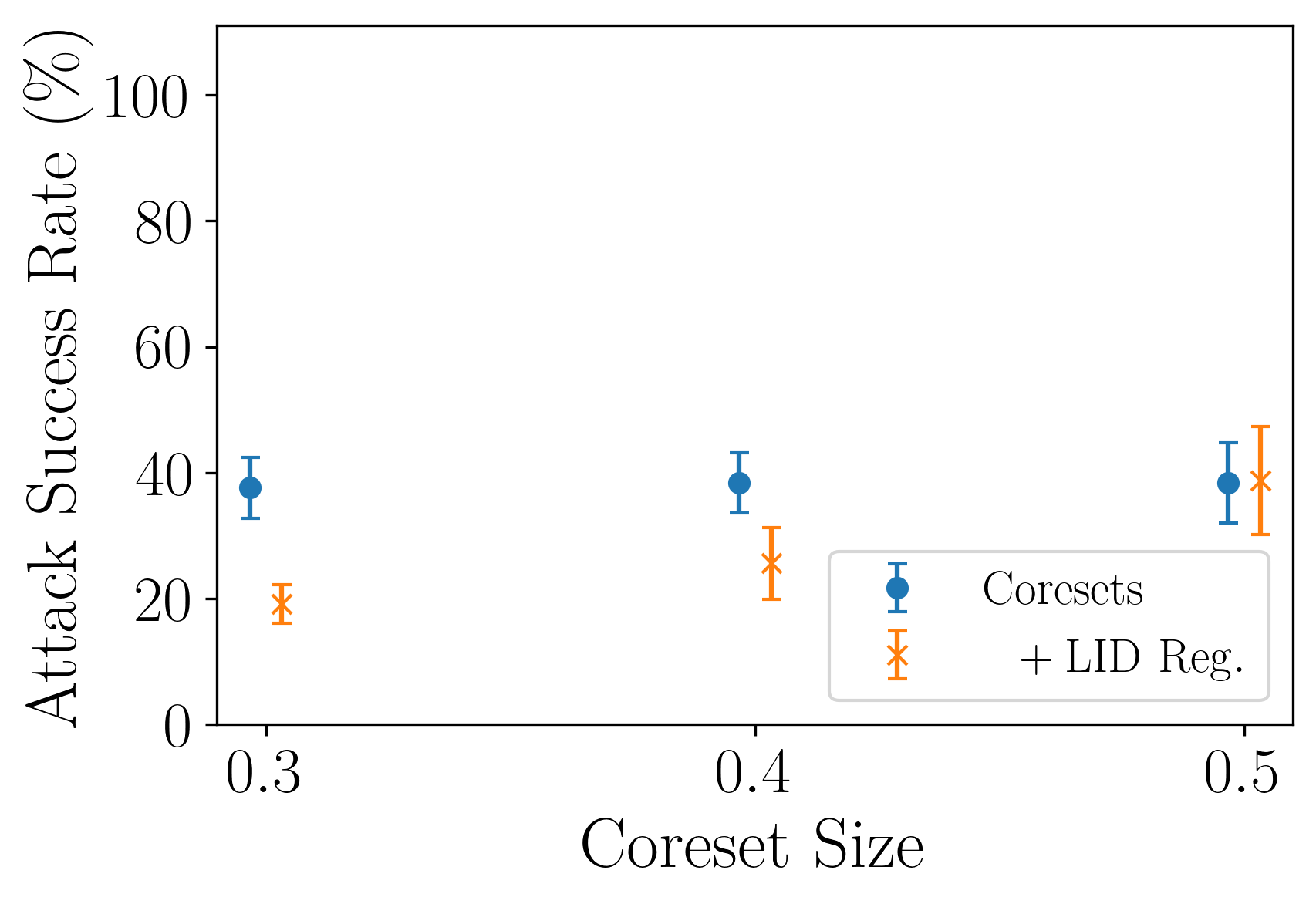}
		\caption{ImageNet-12, HTBA Triggers}
	\end{subfigure}
	\caption{Test accuracy and attack success rate trade-off as the coreset size is increased.}
	\label{fig:ap:coreset_size_trade-off}
	\vskip -0.1in
\end{figure}

\paragraph{Coreset Size Effects.}
\looseness=-1
\Cref{fig:ap:coreset_size} shows the recorded performance measures throughout the training for basic coreset selection~(\Cref{eq:basic_coreset}) and \textsc{Collider}.
In each case, we vary the coreset size to see how the underlying framework changes its behavior.
As expected, by reducing the coreset size we will have cleaner coresets, and hence, get lower attack success rate.
However, there is a trade-off between the coreset size and the clean accuracy, where by reducing the coreset size the validation accuracy also drops.
Another important insight that can be taken from \Cref{fig:ap:coreset_size} is that the LID regularization almost closes the gap in terms of the filtered poison data between different coreset sizes.
Furthermore, it can be seen that the LID regularization is crucial to reduce the attack success rate.
Finally, \Cref{fig:ap:coreset_size_trade-off} also shows the aforementioned trade-off between the clean test accuracy and the attack success rate as the coreset size is increased.
In almost all the cases, \textsc{Collider} results in a more robust neural network.

\paragraph{Number of Nearest Neighbors in LID Computation.}
Finally, we study the effect of the number of neighbors in the LID regularizer.
As pointed out in \Cref{sec:background}, to compute the LID maximum likelihood estimation we need to specify the number of nearest neighbors for each data sample.
This number has a direct relationship with the quality of our estimates.
In particular, the higher the number of neighbors, the more exact the LID estimate.
As such, we expect that as we increase the number of nearest neighbors, \textsc{Collider} would perform better.
However, we cannot increase this value indefinitely as we have limited computational resources.
\Cref{tab:NN} shows the results of our experiments on CIFAR-10 dataset poisoned with checkerboard triggers.
As seen, by increasing the number of nearest neighbors we can improve the neural network robustness against backdoor attacks.

\begin{table*}[htp]\setlength{\tabcolsep}{3.5pt}
	\caption{The effect of the number of nearest neighbors in LID computation on the clean test accuracy (ACC) and attack success rate (ASR) in \% for BadNet data poisoning on CIFAR-10 dataset.
		The poisoned data injection rate is 10\%.}
	\vskip -0.1in
	\label{tab:NN}
	\begin{center}
		\begin{scriptsize}
			\begin{tabular}{ccccccc}
				\toprule
				Nearest Neighb.    & \multicolumn{2}{c}{20}                    & \multicolumn{2}{c}{60}                & \multicolumn{2}{c}{95}\\
				\cmidrule(lr){2-3}                          \cmidrule(lr){4-5}                      \cmidrule(lr){6-7}
				Coreset Size       & ACC               & ASR                   & ACC               & ASR                & ACC               & ASR\\
				\midrule
				0.2                & $76.53 \pm 1.74$	& $2.99 \pm 0.74$      & $75.91 \pm 0.72$	& $4.22 \pm 0.54$   & $75.51 \pm 1.47$  & $4.04 \pm 0.93$\\
				0.3                & $83.44 \pm 0.66$	& $14.57 \pm 7.67$     & $80.66 \pm 0.95$	& $4.80 \pm 1.49$   & $80.94 \pm 0.43$  & $5.02 \pm 1.55$\\
				0.4                & $86.37 \pm 0.30$	& $25.60 \pm 7.90$     & $85.31 \pm 0.19$	& $14.24 \pm 3.11$  & $83.76 \pm 0.58$  & $6.05 \pm 1.73$\\
				0.5                & $88.13 \pm 0.27$	& $67.01 \pm 15.54$    & $87.37 \pm 0.60$	& $15.20 \pm 6.45$  & $86.49 \pm 0.47$  & $8.94 \pm 3.85$\\
				\bottomrule
			\end{tabular}
		\end{scriptsize}
	\end{center}
	\vskip -0.15in
\end{table*}

\begin{figure*}
\centering
\includegraphics[width=0.5\linewidth]{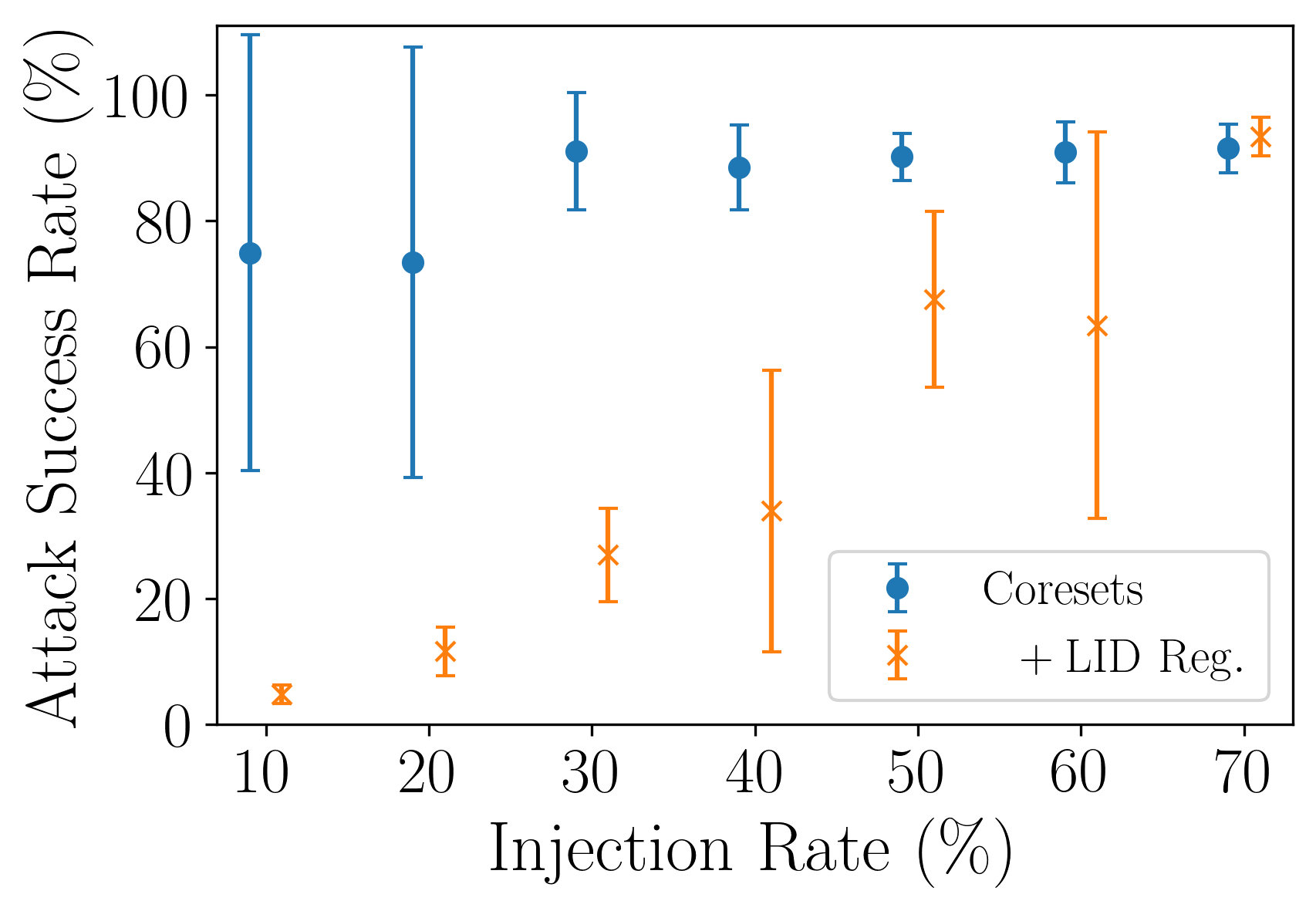}
\caption{Effect of increasing the injection rate on the attack success rate. The coreset size is 0.3.}
\label{fig:ap:asr_vs_inj}
\vskip -0.12in
\end{figure*}

\end{document}